\documentclass{article} 
\usepackage{iclr2026_conference,times}

\usepackage{amsmath,amsfonts,bm}

\def\eqref#1{equation~\ref{#1}}

\def\1{\bm{1}}

\DeclareMathAlphabet{\mathsfit}{\encodingdefault}{\sfdefault}{m}{sl}
\SetMathAlphabet{\mathsfit}{bold}{\encodingdefault}{\sfdefault}{bx}{n}

\usepackage{booktabs}       
\usepackage{amsfonts}       
\usepackage{nicefrac}       
\usepackage{microtype}      

\usepackage{multirow}
\usepackage{makecell}
\usepackage{hyperref}
\usepackage{url}
\usepackage{colortbl}
\usepackage{subcaption}
\usepackage{adjustbox}
\usepackage{pifont}
\usepackage{tabularx}
\usepackage{adjustbox}
\usepackage{wrapfig}
\usepackage{array}
\usepackage{algorithm}
\usepackage{algorithmic}
\usepackage{amsmath}
\usepackage{tcolorbox}
\usepackage{colortbl}
\usepackage{geometry}
\usepackage{graphicx}

\definecolor{dark_red}{RGB}{192,0,0}
\newcommand{\data}{\texttt{Lens}}
\title{LENS: Multi-level Evaluation of Multimodal \\Reasoning with Large Language Models}
\newcommand{\tabincell}[2]{\begin{tabular}{@{}#1@{}}#2\end{tabular}}

\author{Ruilin Yao$^{1,3}$, Bo Zhang$^{1}$, Jirui Huang$^{1,3}$, Xinwei Long$^{2}$, Yifang Zhang$^4$, \\
\textbf{Tianyu Zou$^{1}$, Shili Xiong$^1$, Yi Rong$^1$, Yufei Wu$^{1}$}, \textbf{Shichao Su$^{1}$, Yifan Xu$^{1}$} \\
\textbf{Wenxi Zeng$^{1}$, Zhaoyu Yang$^1$, Guoyou Li$^1$, Shilan Zhang$^1$}, \textbf{Zichan Li$^1$,} \\ 
\textbf{Yaxiong Chen$^{1}$, Shengwu Xiong$^{1}$\footnotemark[1], Peng Xu$^2$\footnotemark[1], Jiajun Zhang$^3$,} \\
\textbf{Bowen Zhou$^{2,4}$, David A. Clifton$^5$, Luc Van Gool$^6$}\\
$^1$ Wuhan University of Technology \quad$^2$ Tsinghua University \quad\\$^3$ Institute of Automation, Chinese Academy of Sciences 
$^4$ Shanghai AI Lab \quad\\
$^5$ University of Oxford \quad$^6$ INSAIT, Sofia Un. St Kliment Ohridski \\
\texttt{xiongsw@whut.edu.cn,} \texttt{peng\_xu@tsinghua.edu.cn}\\
}

\iclrfinalcopy
\begin{document}

\maketitle
\renewcommand{\thefootnote}{\fnsymbol{footnote}}
\footnotetext[1]{Corresponding authors.}
\renewcommand*{\thefootnote}

\begin{abstract}
	Multimodal Large Language Models (MLLMs) have achieved significant advances in integrating visual and linguistic information, yet their ability to reason about complex and real-world scenarios remains limited.
	Existing benchmarks are usually constructed in a task-oriented manner, without a guarantee that different task samples come from the same data distribution. Therefore, 
	they \textcolor{black}{often fall short in evaluating the {synergistic effects} of lower-level perceptual capabilities on higher-order reasoning.}
	To lift this limitation, we contribute \data{}, a multi-\underline{l}evel \underline{e}valuatio\underline{n} benchmark of multimodal rea\underline{s}oning with 3.4K contemporary images and 60K+ human-authored questions covering eight tasks and 12 daily scenarios, forming three progressive task tiers, \textit{i.e.}, perception, understanding, and reasoning.
	One feature is that each image is equipped with rich annotations for all tasks.
	Thus, this data set intrinsically supports evaluating MLLMs 
	to handle image-invariable prompts,
	from basic perception to compositional reasoning.
	In addition, our images have been   collected manually from social media, with $53\%$ published after Jan. 2025.
	We evaluate 15+ frontier MLLMs such as  Qwen2.5-VL,  InternVL3,  GPT-4o  and two reasoning models  QVQ-Max and  Kimi-VL. 
	Most models were released in 2025, and none of them achieve an accuracy beyond $60\%$ in the reasoning tasks. Furthermore, we propose the Self-Driven Multi-Expert Collaborative Framework (SMEC), a framework designed for MLLMs that simulates a panel of experts discussing and exchanging viewpoints via self-generated role-specific prompts. 
    The experimental results confirm the existence of synergistic effects in a hierarchical task structure, where low-level tasks facilitate the reasoning of MLLMs on more complex, high-level tasks.
    Statistical analysis and ablation studies further demonstrate the comprehensiveness of our dataset and the superiority of our methodology. Project page: 
		\url{https://github.com/Lens4MLLMs/lens}. 
            We conducted the ICCV 2025 MARS2 Multimodal Reasoning Challenge on \data{}  
            \url{https://mars2workshop.github.io/iccv2025/}
\end{abstract}
\vspace{-5pt}
\section{Introduction}
Multimodal Large Language Models (MLLMs) have emerged as a rapidly advancing field in artificial intelligence, demonstrating substantial improvements in visual content recognition and multimodal reasoning \citep{zhu2025internvl3, bai2025qwen2,wu2024deepseek,team2024gemini,team2025gemma}. 
Despite their promising capabilities, MLLMs continue to face significant challenges in interpreting complex and real-world visual environments that are inherently dynamic, diverse, and grounded in physicality. However, existing benchmarks remain limited in their ability to evaluate
\textcolor{black}{multi-level reasoning}.

Early evaluations were largely based on classical computer vision tasks \citep{everingham2010pascal,lin2014microsoft, yu2016modeling} and their integration with natural language. The real-world knowledge was often superficial, resulting in weak alignment between visual input and linguistic output. Secondly, these benchmarks are typically constructed under closed-world assumptions, lacking the inter-task consistency needed to assess reasoning across modalities \citep{fu2024mme,li2024survey}. As a result, the absence of quantitative multi-level evaluation hinders meaningful comparison across MLLMs.

\begin{wrapfigure}{t}{0.46\textwidth}\vspace{-2pt}
	\begin{center}
		\includegraphics[width=0.46\textwidth]{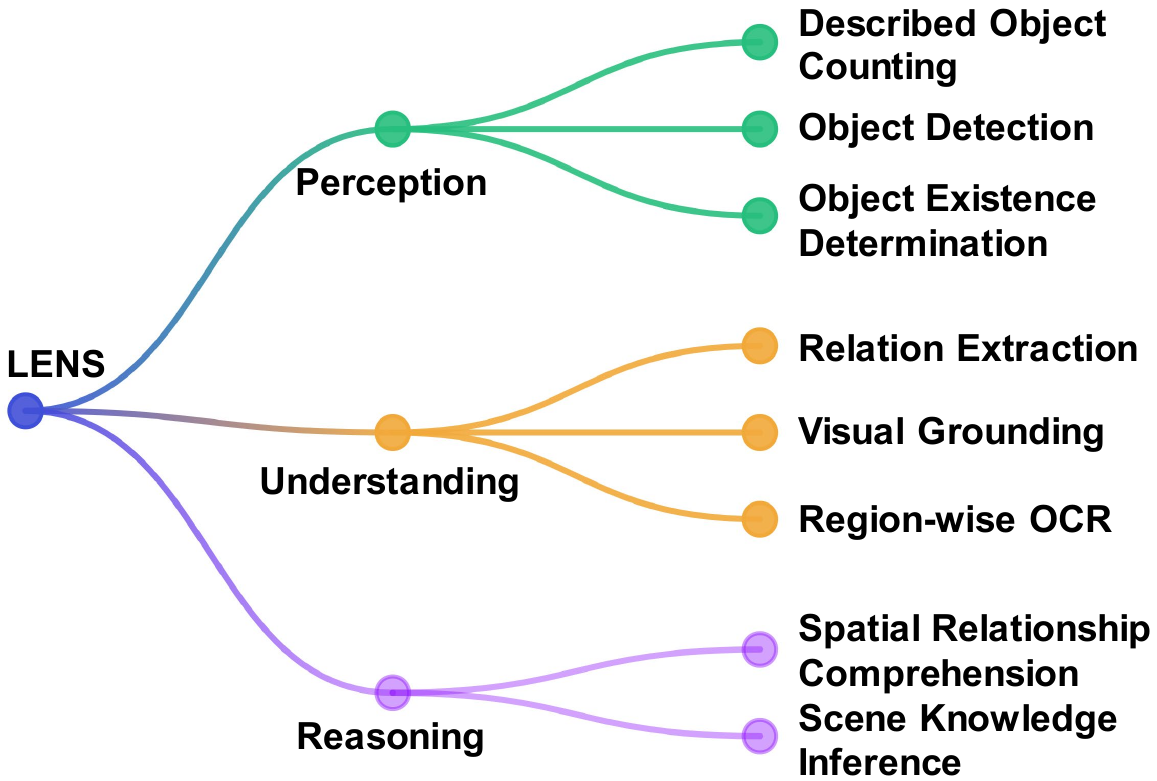}
	\end{center}
	\vspace{-3pt}
	\caption{Illustration of the task split in \data{}.}
	\label{fig:tree_plot}\vspace{-12pt}
\end{wrapfigure}
More recent benchmarks have begun to shift toward open-world evaluation and multimodal reasoning tasks \citep{wu2024v,zang2025contextual}. While this represents progress, current
benchmarks do not adequately assess the nuanced performance necessary to evaluate MLLMs’ progression towards human-like intelligence in real-world settings. 
They require largely primary visual comprehension and fall short of measuring higher-order reasoning and spatial understanding \citep{yue2024mmmu, liu2024mmbench}. Furthermore, data distributions often differed between tasks, so that high performance in perceptual tasks did not necessarily translate into strong inference capabilities in more complex integrated multimodal tasks \citep{fan2025scaling}.  
As a result, they ignore the synergistic effect of the combinations of lower-order perceptual abilities on higher-order reasoning and are hard to provide a fine-grained assessment.

\begin{figure}[t!]
	\centering
	\includegraphics[width=0.95\linewidth]{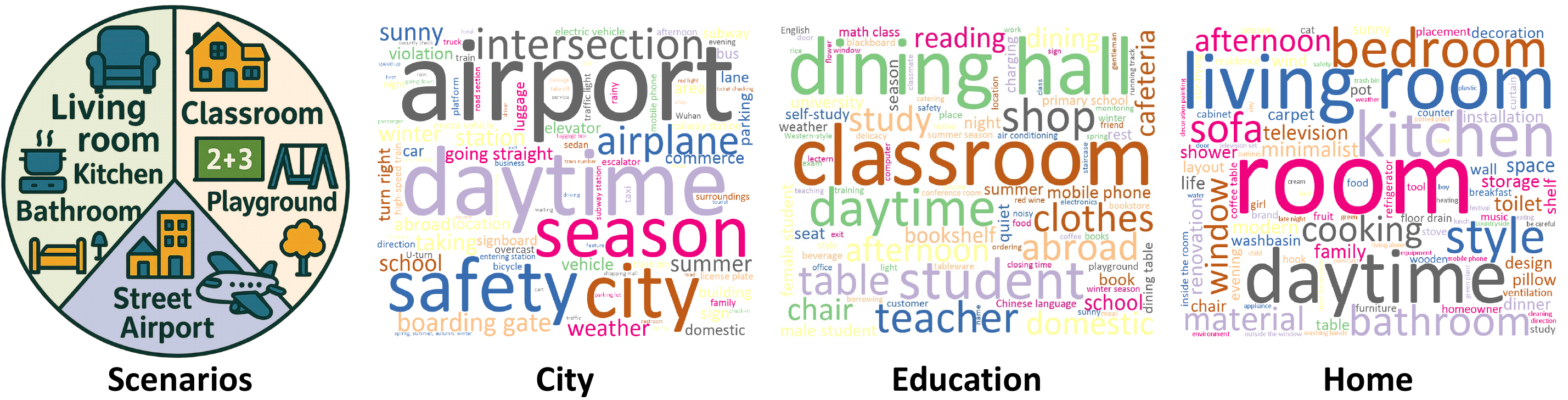}
	\caption{Three core themes, ``Education'', ``City'', and ``Home'', along with their word clouds of the scenario distributions by name size.
	}
	\label{fig:scene}\vspace{-3pt}
\end{figure}

In this study, we propose a hierarchical and comprehensive evaluation framework \data{} specifically designed to assess the multimodal capabilities in real-world scenarios. Our benchmark focuses on both isolated tasks and the integration of perception, understanding, and reasoning—three core tiers essential for intelligent multimodal systems. As shown in Figure \ref{fig:tree_plot}, \data{} encompasses eight tasks, systematically organized into three hierarchical tiers with eight subtasks, and it comprises 3.4K real-world photographs and 60K+ human-authored questions, in 12 diverse scenarios—including streets, stations, schools, homes, and more, which can be roughly divided into three themes: ``Home'', ``Education'',  and ``City'', and we visualize the high-frequency words under different themes in Figure \ref{fig:scene}. 53\% of the images are from 2025 and more than 80\% of the images are from after September 2024, ensuring the content reflects contemporary environments.

For task design, \data{} adopts an open-set configuration, allowing queries to be posed in natural language and grounded in authentic photographic content. This design enables evaluation of model performance in complex, ambiguous, and information-rich settings, better aligning with real-time human demands. Moreover, our benchmark introduces multi-level tasks, which are unified by shared visual contexts, 
making \data{} well-suited for assessing the synergistic effects of lower-level perceptual
abilities (\textit{e.g.}, object detection, localization) on higher-order reasoning tasks. To succeed in \data{}, models must jointly process multimodal input, recall domain knowledge, and conduct multi-step reasoning to arrive at valid conclusions. Our experimental results confirm that current state-of-the-art MLLMs still struggle with these reasoning-heavy tasks, revealing a significant gap between perception and functional understanding. 

To bridge this gap, we propose the Self-Driven Multi-Expert Collaborative Framework (SMEC), a novel reasoning framework that leverages the MLLM itself as a set of specialized experts instantiated through self-generated prompts. 
Unlike tool-calling approaches \citep{wang2025toolgen, gao2025multimodal, liu2025toolace, zhang2024vipact} that rely on external modules, SMEC treats the base MLLM as a versatile reasoning engine: it simulates diverse expert perspectives (\textit{e.g.}, spatial analyst, text interpreter, commonsense reasoner) via role-specific prompts and composes their insights into coherent final answers. This collaborative mechanism encourages the model to extract, expand, and integrate rich, task-relevant information. Our experiments demonstrate that SMEC significantly boosts performance on reasoning tasks within \data{}, validating its potential as a general-purpose, language-native method for enhancing multimodal reasoning. \textbf{We take data privacy, copyright compliance, and platform terms of service seriously. Rigorous collection, filtering, and documentation procedures were implemented and detailed in Section \ref{sec:dataset_collection} and \ref{sec:Ethics}, and Appendix \ref{sec:Privacy}.} To our knowledge, we make the following contributions:
\begin{itemize}
	\setlength{\itemsep}{0.03em}
	\item \textbf{Realistic and Up-to-Date Evaluation.} By leveraging a newly collected set of high-resolution, naturalistic images, our benchmark evaluates the latest multimodal reasoning models in settings that closely reflect real-world complexity.
	\item \textbf{Multi-Level Evaluation.} It supports fine-grained and interpretable evaluation across three core dimensions—perception, understanding, and reasoning—providing a comprehensive view of a model's multimodal competence.
	\item \textbf{Synergistic Capability Evaluation.} Unlike existing benchmarks that often assess tasks in isolation, our framework emphasizes the synergistic effects of lower-level perceptual abilities on higher-order reasoning tasks. The experimental results also confirm that low-level tasks facilitate the reasoning of MLLMs on more complex, high-level tasks (\textit{e.g.}, Scene Knowledge Inference).
	\item \textbf{Towards Generalizable Intelligence.} By capturing both perceptual and reasoning performance in integrated tasks, our benchmark helps identify the gaps between current model capabilities and the requirements of human-aligned reasoning systems and measure the shortcomings of current models.
	\item \textbf{Self-driven Reasoning Enhancement.} We introduce SMEC, a self-driven multi-expert collaborative framework that simulates specialized experts within a single MLLM through self-generated prompts. Unlike tool-calling approaches, SMEC enables modular, multi-perspective reasoning natively, leading to significant gains on complex reasoning tasks.
\end{itemize}
\textbf{Comparison with existing benchmarks.} Compared with existing multimodal benchmarks \citep{liu2024mmbench, yue2024mmmu, li2024embodied}, \data{} provides more contemporary, diverse, and densely annotated visual content. Our benchmark is constructed from contemporary social-media images, ensuring strong timeliness and significantly reducing the risk of contamination from pre-training corpora. In contrast to task-specific datasets \citep{liu2024ocrbench, liu2024finecops, wei2024large}, our benchmark provides rich, multi-task annotations with the same visual content, across perception, understanding, and reasoning, enabling controlled analysis of cross-task synergies within a unified distribution. Additionally, \data{} offers the detailed thought process in real-world reasoning tasks for potential future research. Appendix \ref{related_work} further discusses related work.
\vspace{-12pt}
\section{\data{} Dataset and Benchmark}\label{sec:dataset_bench}
\vspace{-3pt}
\subsection{Data Collection} \label{sec:dataset_collection}
\vspace{-3pt}
The image data collection in our benchmark focuses on real-world scenes to ensure diversity, representativeness, and practicality for visual perception, understanding, and reasoning tasks. To this end, we first defined a set of common real-life scenarios that are highly relevant to typical human visual experiences. The selection principle was that each visual scene should contain distinguishable and representative semantic content. For example, street scenes are usually populated with cars, pedestrians, and storefronts, while indoor environments like classrooms often involve students, teachers, and educational materials. To avoid regional or cultural bias and ensure a broad distribution of content, we collected images from multiple social media platforms, including X (formerly Twitter), Instagram, Weibo, and RedNote. These platforms were chosen due to their global user bases and diverse content coverage across regions and lifestyles. During the collection process, we strictly complied with the copyright and licensing regulations of each platform, ensuring that data was collected only from publicly accessible posts and that no images were downloaded from sources explicitly prohibiting data reuse or redistribution. Moreover, to facilitate the evaluation of multiple subtasks within the same image (\textit{e.g.}, detection, OCR, scene knowledge inference), we curated images that exhibit rich semantic content while maintaining scene clarity. Complex or ambiguous images were manually filtered out to avoid introducing noise that could hinder benchmarking or evaluation consistency. Please note that we manually collect these data that are completely open to the Internet and have complied with the developer agreement of the relevant platform (\textit{e.g.}, Developer Policy of X \footnote{https://developer.x.com/en/use-cases/do-research/academic-research} and Meta \footnote{https://developers.facebook.com/docs/instagram-platform}), ensuring non-commercial use, erasing geographic information, user personal information, etc. from the original data. For further details, please refer to the appendix \ref{sec:Privacy}.
\begin{figure}[t]
	\centering
	\includegraphics[width=0.99\linewidth]{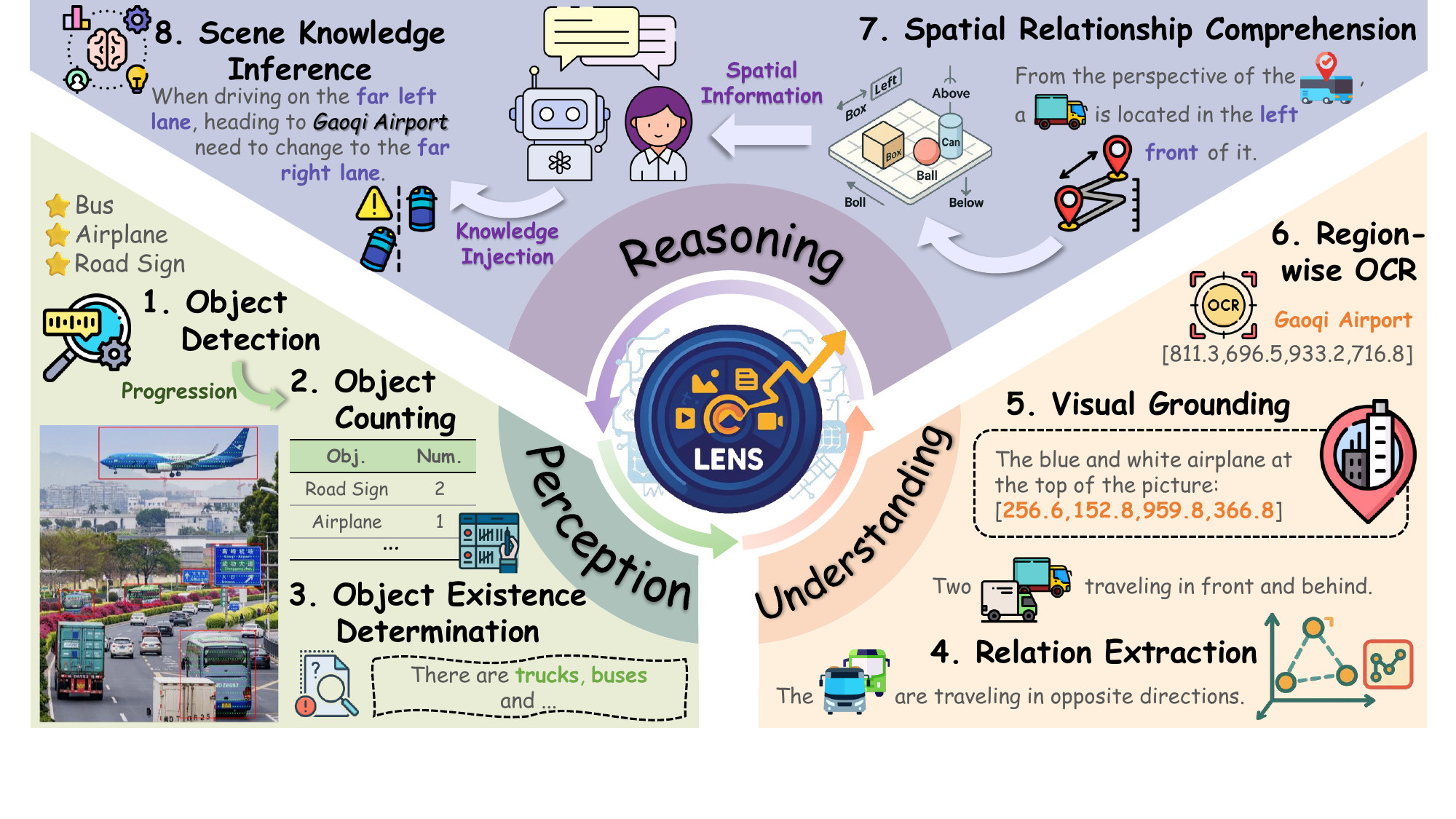}\vspace{-20pt}
	\caption{
		\data{} consists of eight tasks at three levels. \textbf{Perception} tasks focus on recognizing object attribute and counting. \textbf{Understanding} tasks emphasizes localization and inter-object relationships with textual information. \textbf{Reasoning} tasks demand the use of external knowledge beyond the visual input and involve multi-step, complex reasoning processes to arrive at the correct answer.
		\label{fig:intro}}\vspace{-3pt}
\end{figure}
\begin{table}[t]
	\centering
	\caption{Comparison with 
		other recently released multimodal benchmarks.
		}\vspace{-3pt}
		\begin{adjustbox}{max width=\textwidth}
			\begin{tabular}{rcccccccc} 
				\toprule
				Benchmarks & Venue & Att. & Cnt & Loc &  Rel & Reasoning & \tabincell{c}{ Interleaved \\ Image-Text}  & Image Source  \\
				\midrule
				V*~\citep{wu2024v}& CVPR'24 & \ding{56} & \ding{56} & \ding{52} &  \ding{52} & \ding{52} & \ding{56}  &SA-1B \citep{kirillov2023segment} \\
				SPEC~\citep{peng2024synthesize}& CVPR'24 & \ding{52} & \ding{52} & \ding{52} &  \ding{56} & \ding{56} & \ding{56}  &Synthesize \\
				MMVP~\citep{tong2024eyes}& CVPR'24 & \ding{52} & \ding{56} & \ding{56} &  \ding{56} & \ding{56} & \ding{56}  &ImageNet \citep{russakovsky2015imagenet}, LAION-5B \citep{schuhmann2022laion}\\
				HaloQuest~\citep{wang2024haloquest}& ECCV'24 & \ding{52} & \ding{56} & \ding{56} &  \ding{52} & \ding{52} & \ding{56}  &Open Images \citep{kuznetsova2020open}\\
				AS-V2~\citep{wang2024all} & ECCV'24 & \ding{52} & \ding{52} & \ding{52} &  \ding{56} & \ding{52} & \ding{56}  &COCO \citep{caesar2018coco}\\
				MMBench~\citep{liu2024mmbench} & ECCV'24 & \ding{52} & \ding{52} & \ding{52} &  \ding{52} & \ding{52} & \ding{56}  &Internet images \\
				HC-RefLoCo~\citep{wei2024large} & NeurIPS'24 & \ding{56} & \ding{56} & \ding{52} &  \ding{52} & \ding{52} & \ding{56} &Multiple existing datasets \\
				Visual CoT~\citep{shao2024visual} & NeurIPS'24 & \ding{52} & \ding{56} & \ding{56} &  \ding{56} & \ding{52} & \ding{56}  &Multiple existing datasets \\
				MC-Bench \citep{xu2024mc} & arXiv'24 & \ding{56} & \ding{56} & \ding{52} &  \ding{56} & \ding{52} & \ding{52}  &Multiple existing datasets, Internet \\
				CODE~\citep{zang2025contextual}& IJCV'25 & \ding{52} & \ding{52} & \ding{52} &  \ding{56} & \ding{56} & \ding{56}  & Flickr30k series \citep{young2014image, plummer2015flickr30k} \\
				ChatterBox \citep{Tian_Ma_Xie_Ye_2025} & AAAI'25 & \ding{52} & \ding{52} & \ding{52} &  \ding{56} & \ding{52} & \ding{56}  &Visual Genome \citep{krishna2017visual} \\
				\midrule
				\data{} & - & \ding{52} & \ding{52} & \ding{52} &  \ding{52} & \ding{52} & \ding{52}  & \tabincell{c}{{Collect manually from social media} \\ {53\% published later than Jan. 2025}} \\
				\bottomrule
				\multicolumn{9}{c}{``Att.'': Attribute; ``Cnt'': Count; ``Loc'': Localization; ``Rel'': Relation}
			\end{tabular}
		\end{adjustbox}
		\label{comparison_table}
	\end{table}
	
	\subsection{Task Design and Annotation Process}\label{task_definition}
	To construct a comprehensive and diverse benchmark, we recruited over 50 undergraduate and graduate students (including authors) as human annotators to assist in the process of question collection and task annotation and paid the corresponding salary. These annotators were carefully trained to ensure high annotation quality and consistency. As shown in Figure \ref{fig:intro}, the generated questions were divided into three major categories: Perception, Understanding, and Reasoning. For Perception and Understanding, they primarily target the model’s ability to perceive visual objects and align them accurately with natural language descriptions. They emphasize fine-grained visual grounding and object recognition rather than abstract reasoning. At last, reasoning-based questions aim to evaluate the model's ability to understand user intent and reason based on external knowledge, commonsense, physical laws, or background information beyond the purely visual content of the image. Based on these assessment dimensions, we compare \data{} with related multimodal benchmarks in Table \ref{comparison_table} and formulate our challenging open-ended, language-driven tasks: Object Counting (OC), Object Detection (OD), Object Existence Determination (OE), Relation Extraction (RE), Visual Grounding (VG), Region-wise OCR (OCR), Spatial Relationship Comprehension (SRC), and Scene Knowledge Inference (SKI). We provide a
	more detailed introduction of these tasks in Appendix \ref{fig:specific_defination}.
    	\begin{figure}[t]
		\centering
		\includegraphics[width=\linewidth]{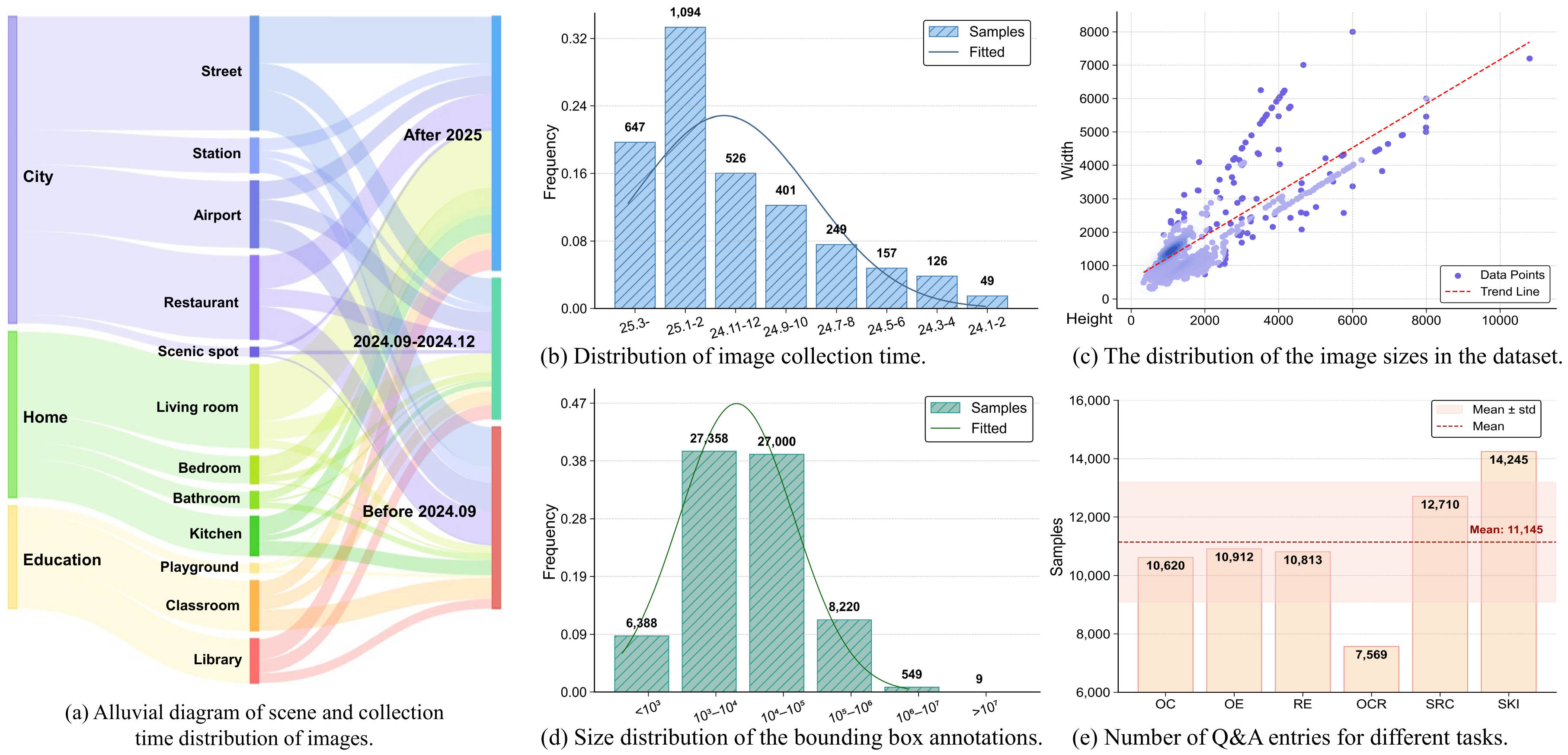}
		\caption{Statistical analysis of our dataset. We visualize the temporal distribution of the images for different scenarios, size distribution of images and bounding box annotations, and number of QA entries for different tasks, demonstrating the timeliness and diversity of our data.
		}
		\label{fig:analysis}
	\end{figure}
	\subsection{Data Analysis }
	We aim to construct a dataset that is not only comprehensive and dynamic but also emphasizes reasoning capabilities. In the following analysis, we demonstrate the strengths of our benchmark in terms of diversity of images and annotations. The quantitative results are visualized in Figure \ref{fig:analysis}.
	
	First, our benchmark incorporates scene-aware content and real-time data. As shown in Figure \ref{fig:analysis}, more than 50\% of the images in our dataset were collected in 2025, and approximately 70\% were collected in November 2024 and beyond, which avoids potential data leakage.
	Many images reflect dynamic scenes (\textit{e.g.}, crowded streets, interactive environments) captured at different times and locations, aligning with real-world scenarios. 
	
	Second, in our dataset, the coverage of a wide range of object categories, scene types, and bounding box annotations further supports diverse downstream tasks from detection to high-level semantic inference and interleaved image-text understanding. As illustrated in Figure \ref{fig:analysis} (c), the high resolution of the images in our dataset makes it challenging for fine-grained understanding of the model and supports evaluation across varying input sizes. Additionally, as shown in Figure \ref{fig:analysis} (d), the various objects are labeled with different sizes of bounding boxes to meet the needs of multi-scale object detection and region-wise OCR evaluation.
	
	Furthermore, beyond perception, our dataset facilitates reasoning-oriented research by supporting tasks that require: Spatial reasoning (\textit{e.g.}, understanding object layouts and geometric relationships). Relational inference (\textit{e.g.}, extracting interactions between objects). Commonsense knowledge application (\textit{e.g.}, inferring the feasibility of a behavior or scene functionalities). Cross-modal alignment (\textit{e.g.}, grounding free-form language to specific visual content). We also analyze the question-answer pairs distribution of these tasks and Figure \ref{fig:analysis} (e) shows that over 60\% of the questions in the dataset go beyond simple recognition, explicitly encouraging models to reason about the scene, context, and user intent. Please refer to the Appendix \ref{low_level} for more low-level visual analysis.
	\begin{figure}[t]
		\centering
		\includegraphics[width=\linewidth]{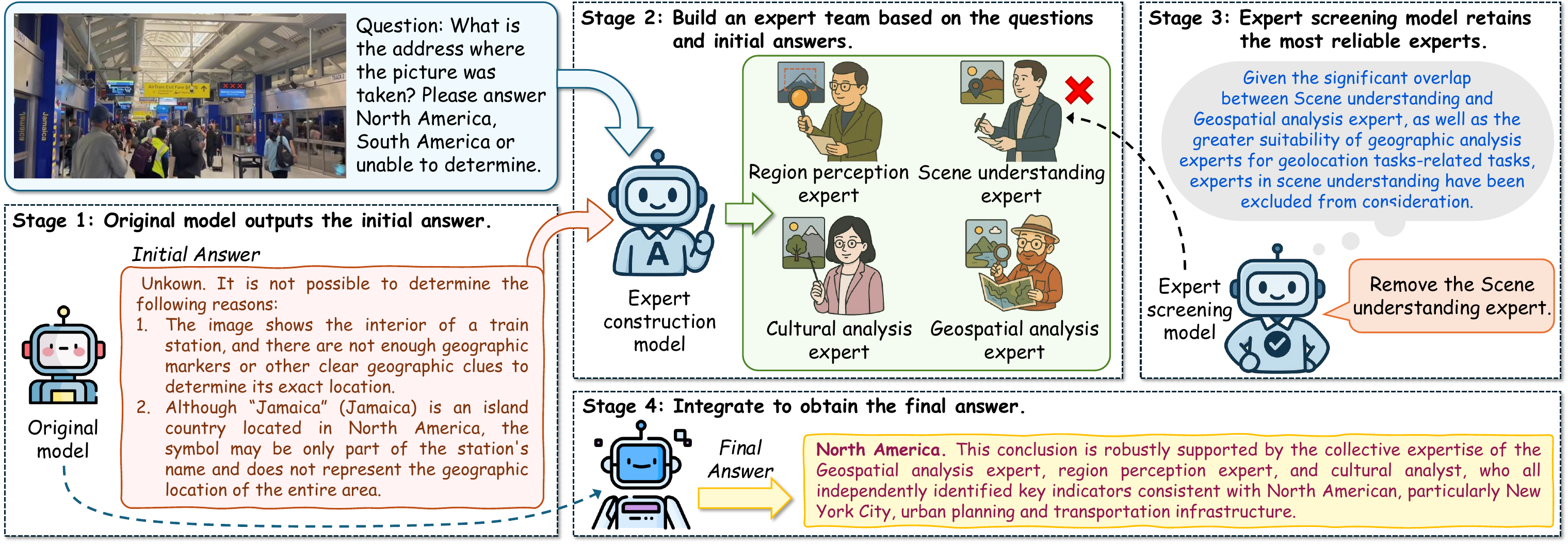}
		\caption{SMEC is a step-wise reasoning framework for answering complex visual questions using the Multi-Expert Collaboration. Starting from an initial model response, SMEC constructs a specialized expert team, to re-evaluate and refine the answer. Through expert screening and integration, unreliable and duplicated experts are filtered out, and a consensus-based final answer is produced, demonstrating the advantages of modular, expert-driven collaboration in visual reasoning tasks.}
		\label{fig:method}
	\end{figure}
	\section{Self-Driven Multi-Expert Collaborative Framework}
	We propose a Self-Driven Multi-Expert Collaborative Framework to tackle complex visual reasoning tasks that require diverse domain expertise and multi-level inference. Built on an instruction-tuned MLLM, our framework dynamically assembles a set of self-generated experts, each embodying a distinct reasoning perspective, as shown in Figure \ref{fig:method}.
	
	\textbf{Expert Generation via Prompted Role Construction.}
	Given a query $q$, the base model $\theta$ first produces a coarse initial answer $a_0$. To enrich the reasoning space, a Meta Generation Prompt $p_g$ is used to iteratively generate expert role descriptions $d_q^t$, simulating specialized agents (\textit{e.g.}, geospatial analyst, cultural analyst). Each valid description yields a new expert response $a_t$, which is added to the answer set $A$. This loop continues until either a diversity criterion is violated or a maximum number of iterations $N_t$ is reached.
	
	\textbf{Prompt Adaptation and Redundancy Filtering.}
	To avoid degenerate expert generation, the framework dynamically updates $p_g$ when  semantically redundant descriptions emerge. This adaptation encourages exploration of novel expert roles. Implicit expert screening is performed by discarding repetitive or low-information descriptions, ensuring a concise yet diverse expert team with minimal computational overhead.
	
	\textbf{Consensus-Driven Answer Integration.}
	The final stage aggregates the expert responses via a Collaboration Prompt $p_c$, prompting $\theta$ to synthesize a unified answer $a_{\text{final}}$ through deliberative reasoning. This mimics human expert panels that reconcile differing views to reach a robust consensus.
	
	We detail a formal description of the process in  Appendix \ref{algorithm1}. In this way, our framework instantiates modular experts purely via prompt-based self-conditioning. Unlike fixed-rule multi-agent systems and tool-calling methods, our framework leverages the generative flexibility of MLLM to dynamically instantiate and evolve its behaviors, without requiring external task-specific supervision.\vspace{-3pt}
    \section{Experiments}\vspace{-3pt}
	\subsection{Evaluation Models}
	To illustrate the difficulty of our benchmark and evaluate the latest advances in current research, we evaluate various MLLMs belonging to three major categories: Closed-source 
	generalist MLLMs, such as GPT-4o \citep{achiam2023gpt} and Gemini2.5 Pro \citep{team2024gemini}. Open-source generalist MLLMs like Qwen2.5-VL \citep{bai2025qwen2}, Deepseek-VL2 \citep{wu2024deepseek}, Gemma3 \citep{team2025gemma}, InternVL3 \citep{zhu2025internvl3}.  Multimodal reasoning models QvQ-Max, Kimi-VL-thinking \citep{team2025kimi} and GLM-4.1V-Thinking \citep{hong2025glm}, focusing on advanced reasoning capabilities. 
	The release dates of these models are distributed from Dec. 2024 to Apr. 2025.
	\subsection{Evaluation Strategy}
	To ensure a fair and efficient assessment of model performance across our benchmark, we adopt two evaluation strategies for main results. For perception and understanding tasks, models were evaluated based on their direct outputs without additional inference-time computations. For complex reasoning tasks, which require deeper multi-step inference, we allow models to generate multiple candidate responses per question and the final prediction is then selected via majority voting \citep{liu2025bag}. 
	For qualitative judgment, we follow prior work \citep{wang2023large} and employ a large language model (\textit{e.g.}, GLM4-flash \citep{glm2024chatglm}) as an automatic evaluator. The LLM is prompted to produce multiple pieces of evaluation evidence for calibration, comparing the model-generated responses against human-annotated answers, aiming to offer a consistent framework for evaluating model performance across diverse tasks.
	
	\subsection{ Evaluation Results}
	\begin{table}[t]\setlength\tabcolsep{3pt}
		\centering
		\caption{Comparison of state-of-the-art methods on \data{}. We evaluate object detection (OD) performance using AP$_{50}$ \citep{lin2014microsoft}, visual grounding (VG) performance with ACC$@0.5$ \citep{xiao2024towards}, and use accuracy for other tasks. Task abbreviations follow the definitions provided in Section \ref{task_definition}.
			``MoE 1B/3B'' denotes 3B Mixture of Experts model with 1B parameters activated. ``N/A'' denotes the official documentation does not confirm that the model is applicable for the task. Best performing models are shaded in \colorbox{red!10}{red}.
		}\vspace{-3pt}
		\resizebox{0.99\textwidth}{!}{
			\begin{tabular}{l r ccc ccc cc}
				\toprule
				\multirow{2}{*}{Methods} & \multirow{2}{*}{\tabincell{c}{Model \\ size}} & \multicolumn{3}{c}{Perception} & \multicolumn{3}{c}{Understanding} & \multicolumn{2}{c}{Reasoning}  \\ 
				\cmidrule(r){3-5} \cmidrule(r){6-8} \cmidrule(r){9-10} 
				& &  OC  & OD & OE&  RE & VG  & OCR & SRC  & SKI   \\
				\midrule
				\multicolumn{10}{l}{\textbf{MLLM (closed source)}} \\ 
				\midrule
				\includegraphics[height=0.9em]{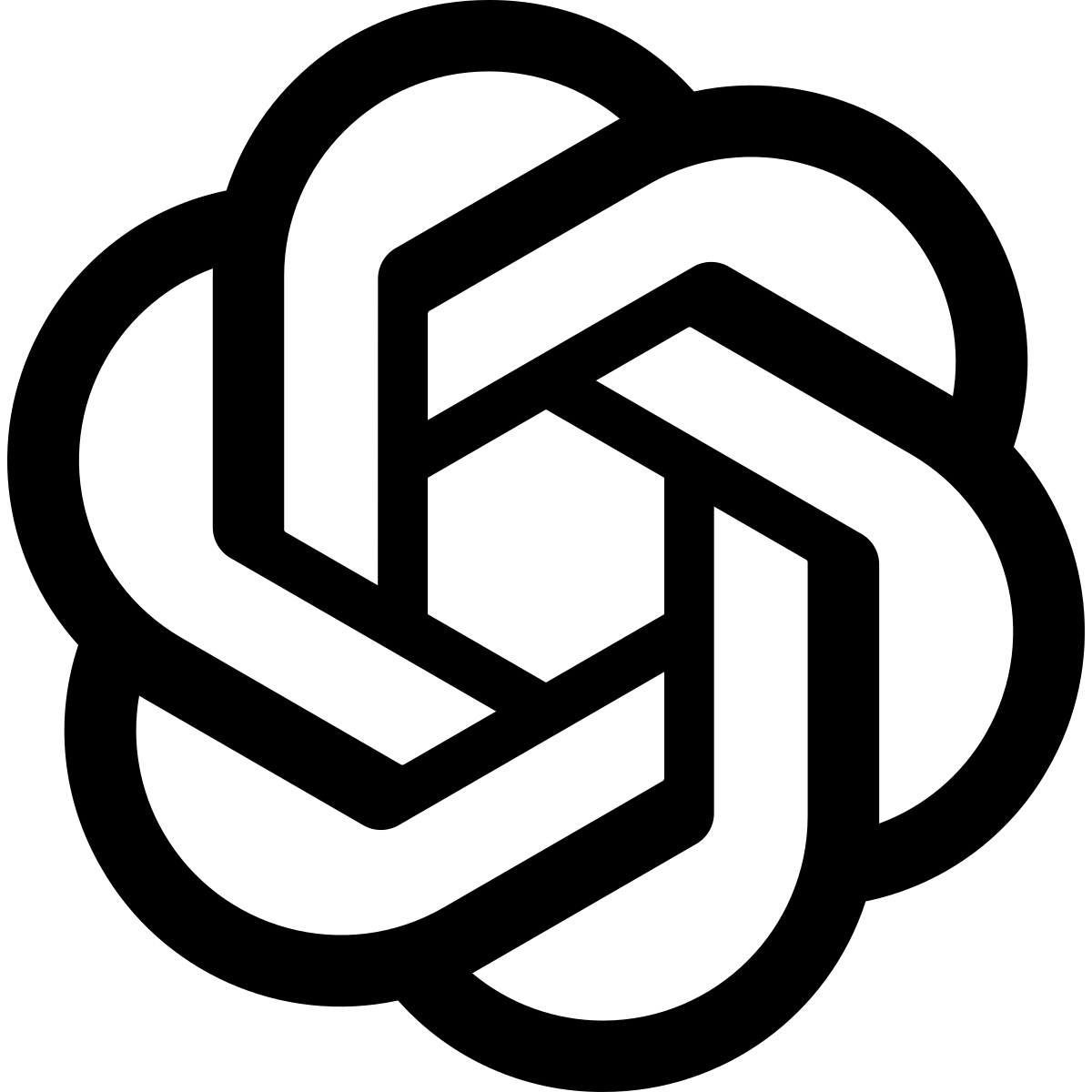} ~GPT-4o & - & 54.32 & N/A & 85.09
				& 72.77 & N/A & 42.86 & 51.14 & 55.20 
				\\
				\includegraphics[height=0.9em]{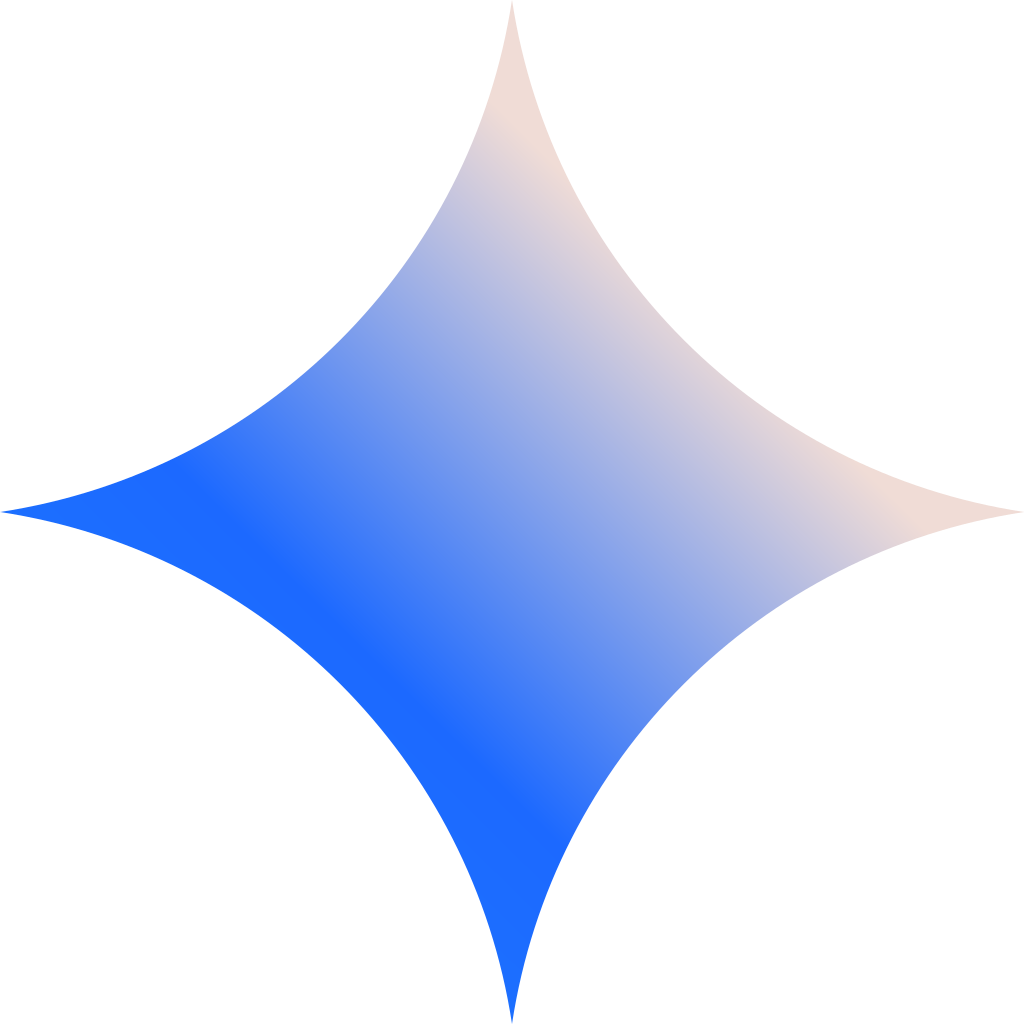} ~Gemini2.5-Pro & - & 60.18 & \cellcolor{red!10} 47.40 & 86.59
				& \cellcolor{red!10} 76.52 & 25.61 & 61.95 & \cellcolor{red!10} 56.20 & \cellcolor{red!10} 59.31  \\
				\midrule
				\multicolumn{10}{l}{\textbf{Open source}} \\ 
				\midrule
				\includegraphics[height=0.9em]{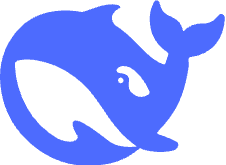} ~Deepseek-VL2-tiny & MoE 1B/3B  & 56.22 & 21.12 & 72.11
				& 58.73 & 16.09 & 44.01 & 38.97 & 45.12  \\

				\includegraphics[height=0.9em]{./logo/deepseek.png} ~Deepseek-VL2  & MoE 4.5B/27B & 61.41 & 46.08 & 77.68
				& 69.18 & 42.47& 48.76 & 44.58 & 49.50  \\
				\includegraphics[height=0.9em]{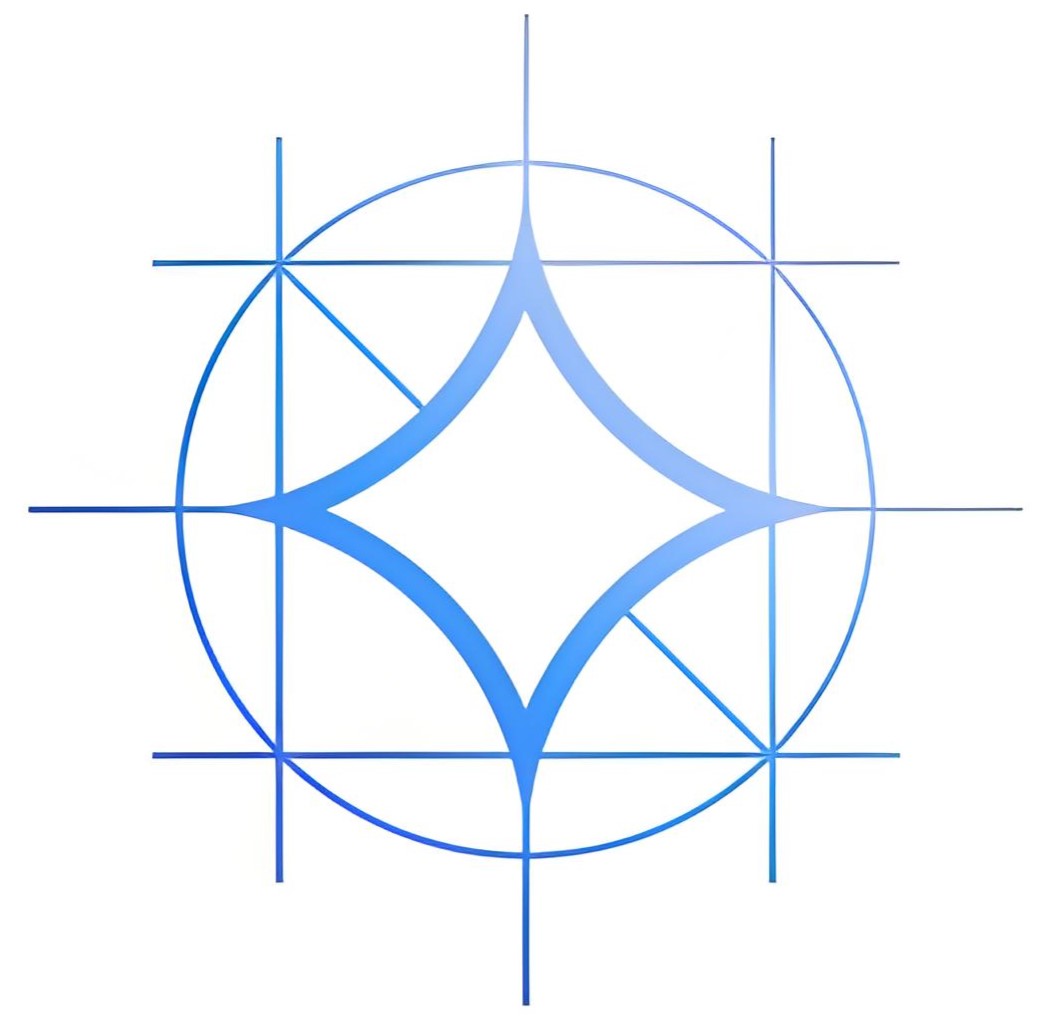} ~Gemma3  & 4B & 38.85 & N/A & 71.88
				& 62.98 & N/A & 27.03 & 39.53 & 45.18 \\
				\includegraphics[height=0.9em]{./logo/gemma3.jpg} ~Gemma3  & 12B & 44.65 & N/A & 73.21
				& 62.78 & N/A & 33.98 & 43.33 & 48.56  \\
				\includegraphics[height=0.9em]{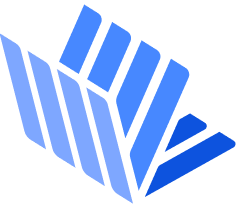} ~InternVL3 &2B& 55.81 & 18.39 & 71.96
				& 64.49 & 15.22 & 45.51 & 40.56 & 48.59  \\
				\includegraphics[height=0.9em]{./logo/InternVL.png} ~InternVL3 &9B& 55.63 & 25.79 & 77.49
				& 67.18 & 18.18 & 48.79 & 44.69 & 51.32  \\
				\includegraphics[height=0.9em]{./logo/InternVL.png} ~InternVL3 &38B& \cellcolor{red!10} 62.78 & 43.44 & 81.60 & 71.37 & 24.98 & 51.72 & 47.18 & 51.85  \\
				\includegraphics[height=0.9em]{./logo/InternVL.png} ~InternVL3 &78B& 61.38 & 47.44 & 84.87 & 74.93 & 27.24 & 54.21 & 49.39 & 55.17  \\
				\includegraphics[height=0.9em]{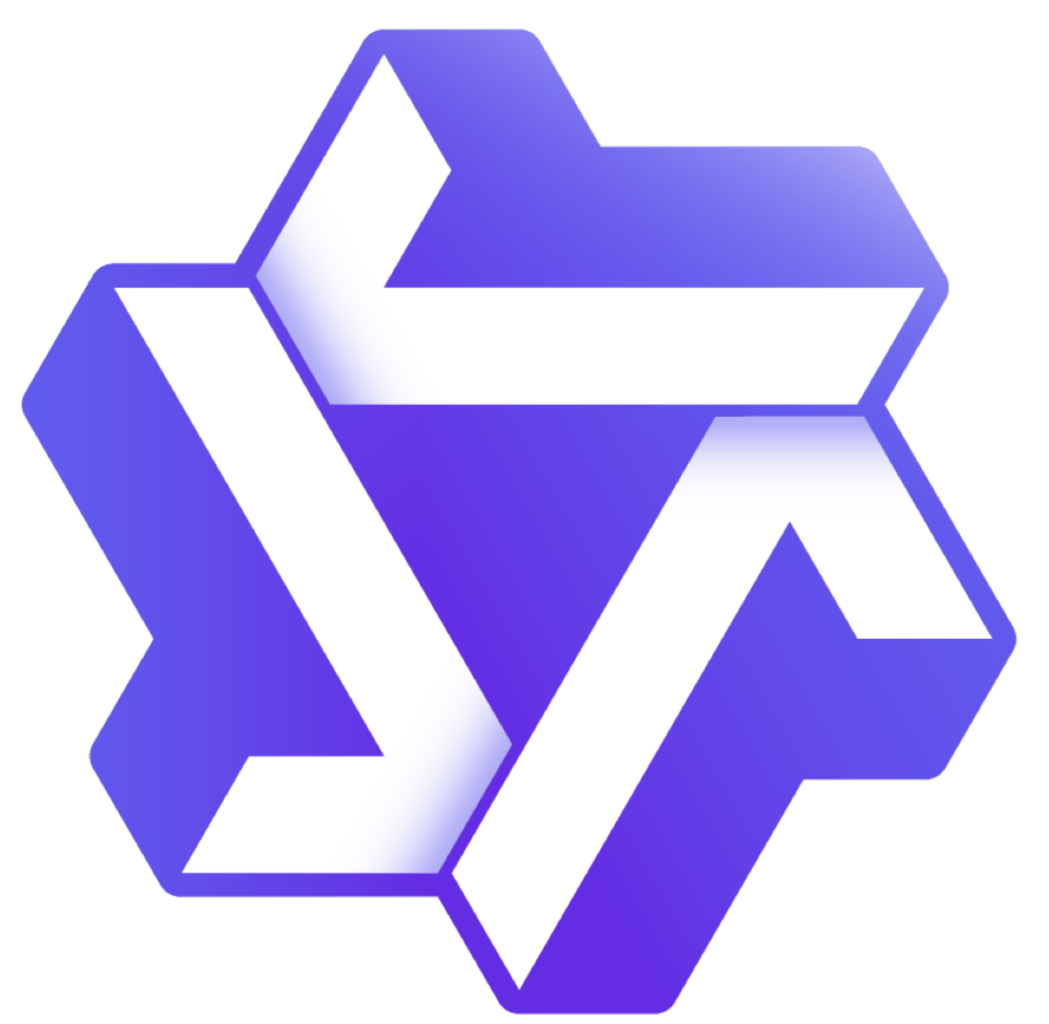} ~Qwen2.5-VL & 3B & 58.76 & 35.16 & 74.01 & 66.52 & 39.44 & 52.43 & 40.33 & 46.50  \\
				\includegraphics[height=0.9em]{./logo/qwen.png} ~Qwen2.5-VL & 7B & 58.35 & 37.75 & 83.75 & 71.58 & 40.11 & 61.65 & 46.28 & 48.87  \\
				\includegraphics[height=0.9em]{./logo/qwen.png} ~Qwen2.5-VL & 32B & 62.25 & 39.93 & 83.60 & 74.57 & 41.15 & 65.64 & 51.66 & 51.54  \\
				\includegraphics[height=0.9em]{./logo/qwen.png} ~Qwen2.5-VL & 72B & 59.75 & 43.48 & \cellcolor{red!10} 85.67 &  75.98 & \cellcolor{red!10} 44.98 & \cellcolor{red!10} 68.51 & 53.65 & 54.79  \\
				\midrule
				\multicolumn{10}{l}{\textbf{Reasoning model}} \\ 
				\midrule
				\includegraphics[height=0.9em]{./logo/qwen.png} ~QVQ-Max & 72B & 49.95 & N/A & 85.37 & 74.01 & N/A & 58.67 & 50.80 & 58.86  \\
				\includegraphics[height=0.9em]{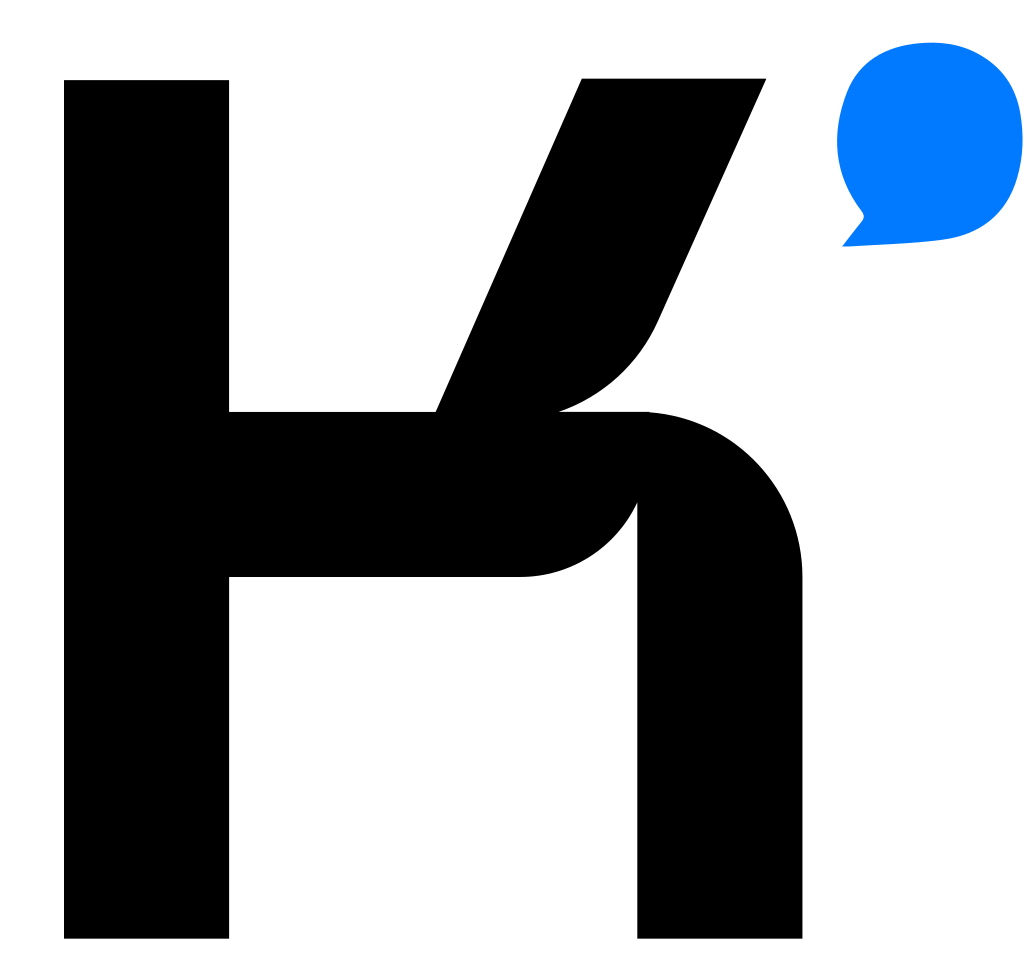} ~Kimi-VL-thinking& MoE 2.8B/16B & 46.87 & N/A & 72.77 & 48.16 & N/A &30.21 & 29.40 & 36.44  \\
				\bottomrule
			\end{tabular}
		}\vspace{-5pt}
		\label{tab:sota_method}
	\end{table}
	We evaluate a suite of state-of-the-art Multimodal Large Language Models on our benchmark, which spans three tiers and eight tasks. Results, as shown in Table \ref{tab:sota_method}, reveal insights into model scaling, inter-task dependencies, and capability gaps in current MLLMs.
	
	\textbf{Model Scaling and General Trends.}
	We observe a consistent performance gain with increased model size in both closed- and open-source models. For example, Qwen2.5-VL improves steadily from 3B to 72B, achieving top performance on reasoning tasks. InternVL3 shows similar gains in OD, rising from 18.39\% (2B) to 47.44\% (78B), though performance saturates at higher scales. These trends confirm that scaling remains a key driver for multimodal reasoning, albeit with diminishing returns in some subtasks.
	
	\textbf{Perception: Foundation for Higher Cognition.}
	Perception-level tasks form the backbone of visual reasoning. Closed-source models like Gemini2.5-Pro and GPT-4o excel at OE (86.59\% and 85.09\%, respectively), although OD support is lacking. Among open-source models, Deepseek-VL2 and Qwen2.5-VL-72B deliver competitive OD and OC performance. Notably, models with stronger perception capabilities tend to exhibit superior reasoning performance, highlighting the foundational role of low-level visual understanding.
	
	\textbf{Understanding: Progress and Bottlenecks.}
	Understanding tasks assess models’ ability to interpret structured visual semantics with textual information. Gemini2.5-Pro leads in RE (76.52\%) and OCR (61.95\%), showcasing robust relational and textual grounding. However, VG remains a bottleneck even for large-scale models like InternVL3-78B (27.24\%) and Qwen2.5-VL-72B (44.98\%), suggesting persistent challenges in fine-grained spatial-semantic alignment.
	
	\textbf{Reasoning: High-Level Generalization.}
	Reasoning tasks are the most demanding. Closed-source models such as GPT-4o and Gemini2.5-Pro achieve strong results (51.14\%/56.20\% on SRC and 55.20\%/59.31\% on SKI). Among open-source models, Qwen2.5-VL-72B leads, while the reasoning-specialized QVQ-Max approaches closed-source performance (58.86\% on SKI) despite lacking OD and VG capabilities. This suggests that explicit reasoning models can partially compensate for perceptual limitations, likely relying on test-time scaling rather than grounded perception.
	\begin{figure}[t]
		\centering
		\includegraphics[width=\linewidth]{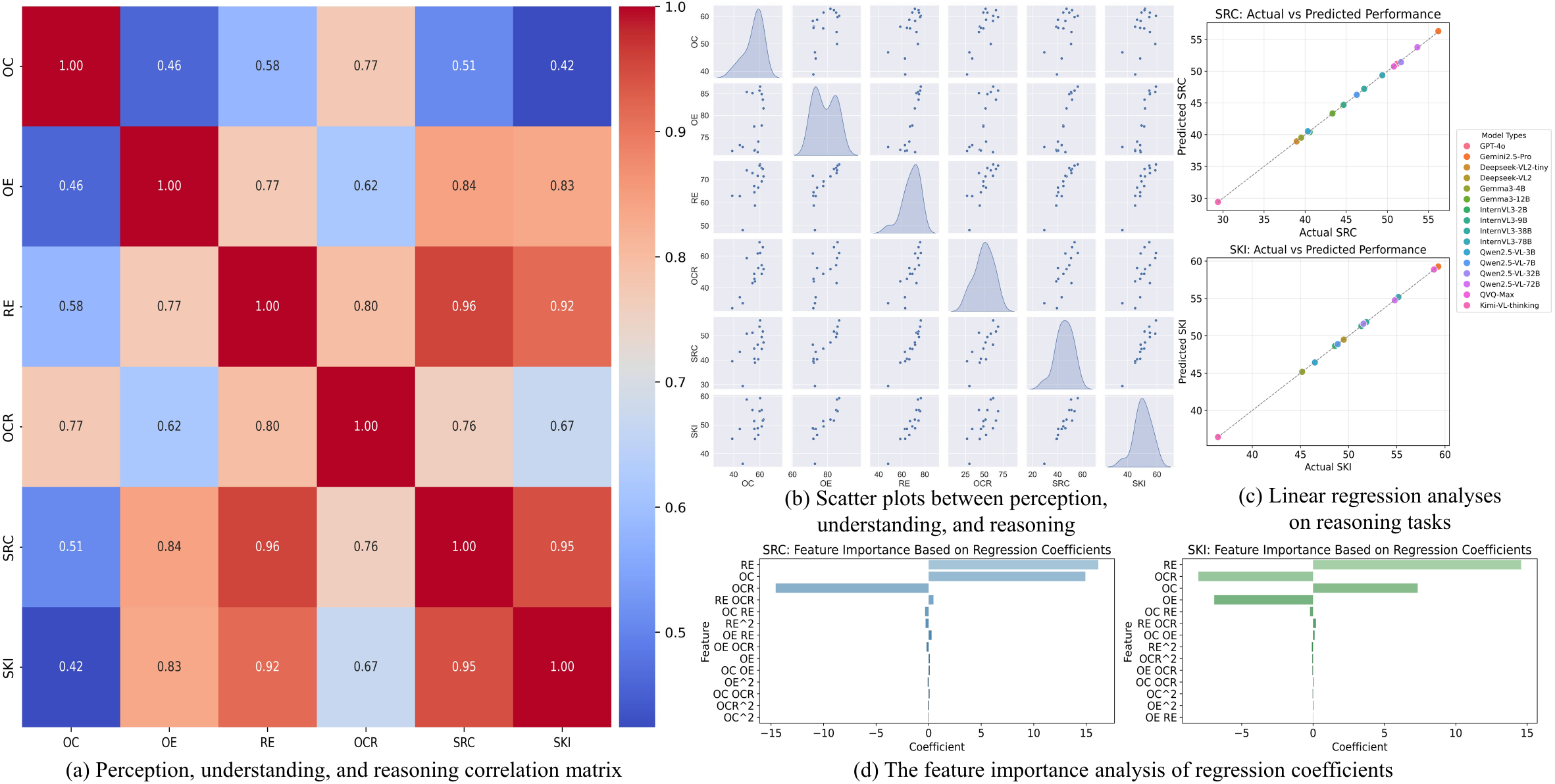}
		\caption{Statistical analysis of model accuracy and synergies between different tasks.}
		\label{fig:fine_grained}\vspace{-2pt}
	\end{figure}
	\subsection{Synergistic effects analysis}
    To analyze the cross-task performance patterns of different models, we perform a statistical analysis of the synergies between different tasks and visualized the results as in Figure \ref{fig:fine_grained}. We compute the Pearson correlation coefficients between Perception and Understanding tasks and observe notable interdependencies. OC and RE exhibit a strong positive correlation of $0.73$, while OE and OCR show a similarly significant correlation of $0.67$. These results indicate that effective performance in perception directly contributes to understanding, which in turn underpins downstream reasoning. Scatter plot visualizations further confirm these links, OCR, in particular, correlates strongly with both SRC and SKI, underscoring its central role in enabling semantic reasoning. Linear regression analyses reinforce these findings: OE and OCR are strong predictors of SRC, while OC and RE significantly influence SKI, highlighting how object-level detection and relational reasoning jointly support high-level inference. Finally, we apply second-order polynomial regression and the feature importance analysis of regression coefficients reveals task-specific contributions. These insights collectively demonstrate the layered structure of visual reasoning pipelines, where perception and understanding stages must be well-aligned to support robust inference. For further analysis, please refer to the appendix \ref{more_synergistic}.
\subsection{Effectiveness of SMEC}
	\begin{wraptable}{t}{0.5\textwidth}\vspace{-16pt}
		\begin{center}
			\caption{Accuracy comparison with different settings.}\label{tab:sota2}\vspace{-5pt}
			\resizebox{0.5\textwidth}{!}{
				\begin{tabular}{l|c|c|c}
					\toprule
					Methods & Model & Iterations & Performance \\
					\midrule
					Direct & Qwen2.5VL-7b & - & 39.80\\
                    Majority voting & Qwen2.5VL-7b & - & 40.66\\
					Self-Refine  & Qwen2.5VL-7b & - & 40.51 \textcolor{dark_red}{(+0.71)}\\
					SMEC & Qwen2.5VL-7b & 1 & 41.35 \textcolor{dark_red}{(+1.55)} \\
					SMEC & Qwen2.5VL-7b & 2 & 42.97 \textcolor{dark_red}{(+3.17)}\\
					SMEC & Qwen2.5VL-7b & 3 & 43.24 \textcolor{dark_red}{(+3.44)}\\
					\midrule
					Direct & Qwen2.5VL-32b & - & 49.17\\
					SMEC & Qwen2.5VL-32b & 3 & 52.44 \textcolor{dark_red}{(+3.27)}\\
                    \midrule
                    Direct (Full data) & Qwen2.5VL-32b & - & 51.54\\
					SMEC (Full data)& Qwen2.5VL-32b & 3 & 54.66 \textcolor{dark_red}{(+3.12)}\\
					\bottomrule
			\end{tabular}}
		\end{center}\vspace{-5pt}
	\end{wraptable} 
	To further evaluate the effectiveness of our proposed SMEC framework, we conduct experiments on the Scene Knowledge Inference (SKI) task using both a sampled subset of 3,500 question–answer pairs and the full test set (Table \ref{tab:sota2}). Compared to the direct inference baseline, SMEC consistently improves performance across both model scales and different iteration depths. For Qwen2.5VL-7B, accuracy increases from 39.80\% to 43.24\% as the number of iterations $N_t$ grows from 1 to 3, demonstrating the benefits of multi-step expert collaboration. Notably, SMEC also outperforms Self-Refine \citep{madaan2023self} and Majority voting \citep{chen2024more} under the same setting. 
    A similar trend appears at larger scales. With Qwen2.5VL-32B, SMEC improves accuracy from 49.17\% to 52.44\% with three iterations, confirming that the benefits of iterative expert collaboration scale with model capacity. Importantly, when evaluated on the full test set rather than the 3,500-sample subset, SMEC continues to provide consistent improvements (+3.12\%). This indicates that SMEC’s gains are not distribution-specific and remain stable under more comprehensive evaluation conditions.
	\section{Conclusion}
	We contribute \data{}, a multi-level benchmark designed to evaluate Multimodal Large Language Models (MLLMs) across perception, understanding, and reasoning. Unlike prior benchmarks, \data{} aligns all tasks to the same set of realistic, contemporary images, enabling fine-grained analysis of how low-level visual capabilities support higher-order reasoning. 
	The evaluation of recent MLLMs further reveals a consistent performance gap in reasoning tasks, highlighting the limitations of current models in integrating perception and cognition. To address this, we proposed SMEC, a self-driven multi-expert collaborative framework that prompts the MLLM to simulate a panel of specialized agents. 
	Together, \data{} and SMEC offer a new paradigm for evaluating and enhancing reasoning intelligence in MLLMs, paving the way to more robust, human-aligned multimodal intelligence.
	\section{Ethics Statement} \label{sec:Ethics}
    \textbf{Data Collection and Privacy.} All images in the \data{} dataset were \textbf{collected manually from publicly available posts} on social media platforms. We respect and strictly complied with the developer agreements and copyright regulations of these platforms. No private or restricted data was accessed, and all collection adhered to academic research policies. This dataset is intended strictly for non-commercial academic research purposes. If any content in this project is found to raise concerns related to privacy, copyright, or legal compliance, please contact us at \textbf{yaoruilin@whut.edu.cn}. We will promptly review the request and are willing to remove, modify, or withdraw the sensitive data or materials.

    \textbf{Human Annotation.} The dataset was constructed with the assistance of more than 50 undergraduate and graduate student annotators, who were trained to ensure annotation quality and consistency. All annotators were fairly compensated for their work. The annotation process involved only task-related labeling (\textit{e.g.}, bounding boxes, question-answer generation) and did not involve collection of personal or sensitive information about the annotators.

    \textbf{Bias, Fairness, and Representation.}
Although images were sourced from global platforms to ensure diversity of cultural and regional content, dataset bias may still exist due to platform-level demographic imbalances and uneven scenario representation. We acknowledge that these limitations may influence model evaluation results, and we encourage future work to further broaden geographic and cultural coverage.

\textbf{Research Integrity and Transparency.}
This study follows established standards of research integrity. We provide detailed dataset descriptions, evaluation protocols, and implementation details. We have no conflicts of interest or external sponsorship that might bias the study.
\section{Reproducibility statement}
 We have taken extensive measures to ensure the reproducibility of our results. A complete description of the \data{} dataset, including data collection principles, filtering steps, and annotation procedures, is provided in Section 2 and Appendix. Details of the evaluation tasks, metrics, and benchmark comparisons are reported in Section 4, with additional analyses in Appendix. The implementation of our proposed Self-Driven Multi-Expert Collaborative Framework (SMEC), including the expert construction, screening, and integration process, is fully described in Section 3 and formalized in Appendix A.9. Hyperparameter settings, model configurations, and ablation experiments are included in the appendix to facilitate replication of results.  Together, these resources ensure that both dataset construction and methodological contributions can be faithfully reproduced by the research community.

 \section*{Acknowledgement}
This work is supported by the National Key Research and Development Program of China
No.2022ZD0160604, and NSFC No.62306162.
	\bibliography{iclr2026_conference}

@article{liu2024ocrbench,
  title={OCRBench: on the hidden mystery of OCR in large multimodal models},
  author={Liu, Yuliang and Li, Zhang and Huang, Mingxin and Yang, Biao and Yu, Wenwen and Li, Chunyuan and Yin, Xu-Cheng and Liu, Cheng-Lin and Jin, Lianwen and Bai, Xiang},
  journal={Science China Information Sciences},
  volume={67},
  number={12},
  pages={220102},
  year={2024},
  publisher={Springer}
}

@inproceedings{yu2016modeling,
  title={Modeling context in referring expressions},
  author={Yu, Licheng and Poirson, Patrick and Yang, Shan and Berg, Alexander C and Berg, Tamara L},
  booktitle={European Conference on Computer Vision},
  pages={69--85},
  year={2016},
  organization={Springer}
}

@article{liu2024finecops,
  title={FineCops-Ref: A new Dataset and Task for Fine-Grained Compositional Referring Expression Comprehension},
  author={Liu, Junzhuo and Yang, Xuzheng and Li, Weiwei and Wang, Peng},
  journal={arXiv preprint arXiv:2409.14750},
  year={2024}
}

@article{wei2024large,
  title={A Large-Scale Human-Centric Benchmark for Referring Expression Comprehension in the LMM Era},
  author={Wei, Fangyun and Zhao, Jinjing and Yan, Kun and Zhang, Hongyang and Xu, Chang},
  journal={Advances in Neural Information Processing Systems},
  volume={37},
  pages={69566--69587},
  year={2024}
}

@article{krishna2017visual,
  title={Visual genome: Connecting language and vision using crowdsourced dense image annotations},
  author={Krishna, Ranjay and Zhu, Yuke and Groth, Oliver and Johnson, Justin and Hata, Kenji and Kravitz, Joshua and Chen, Stephanie and Kalantidis, Yannis and Li, Li-Jia and Shamma, David A and others},
  journal={International journal of computer vision},
  volume={123},
  pages={32--73},
  year={2017},
  publisher={Springer}
}

@article{cheng2024emotion,
  title={Emotion-llama: Multimodal emotion recognition and reasoning with instruction tuning},
  author={Cheng, Zebang and Cheng, Zhi-Qi and He, Jun-Yan and Wang, Kai and Lin, Yuxiang and Lian, Zheng and Peng, Xiaojiang and Hauptmann, Alexander},
  journal={Advances in Neural Information Processing Systems},
  volume={37},
  pages={110805--110853},
  year={2024}
}

@inproceedings{rachabatuni2024context,
  title={Context-aware chatbot using MLLMs for Cultural Heritage},
  author={Rachabatuni, Pavan Kartheek and Principi, Filippo and Mazzanti, Paolo and Bertini, Marco},
  booktitle={Proceedings of the 15th ACM Multimedia Systems Conference},
  pages={459--463},
  year={2024}
}

@article{liu2023multilingual,
  title={Are multilingual llms culturally-diverse reasoners? an investigation into multicultural proverbs and sayings},
  author={Liu, Chen Cecilia and Koto, Fajri and Baldwin, Timothy and Gurevych, Iryna},
  journal={arXiv preprint arXiv:2309.08591},
  year={2023}
}

@article{dong2022survey,
  title={A survey on in-context learning},
  author={Dong, Qingxiu and Li, Lei and Dai, Damai and Zheng, Ce and Ma, Jingyuan and Li, Rui and Xia, Heming and Xu, Jingjing and Wu, Zhiyong and Liu, Tianyu and others},
  journal={arXiv preprint arXiv:2301.00234},
  year={2022}
}

@article{wei2022chain,
  title={Chain-of-thought prompting elicits reasoning in large language models},
  author={Wei, Jason and Wang, Xuezhi and Schuurmans, Dale and Bosma, Maarten and Xia, Fei and Chi, Ed and Le, Quoc V and Zhou, Denny and others},
  journal={Advances in neural information processing systems},
  volume={35},
  pages={24824--24837},
  year={2022}
}

@article{alayrac2022flamingo,
  title={Flamingo: a visual language model for few-shot learning},
  author={Alayrac, Jean-Baptiste and Donahue, Jeff and Luc, Pauline and Miech, Antoine and Barr, Iain and Hasson, Yana and Lenc, Karel and Mensch, Arthur and Millican, Katherine and Reynolds, Malcolm and others},
  journal={Advances in neural information processing systems},
  volume={35},
  pages={23716--23736},
  year={2022}
}

@inproceedings{yao2024visual,
  title={Visual Grounding with Multi-modal Conditional Adaptation},
  author={Yao, Ruilin and Xiong, Shengwu and Zhao, Yichen and Rong, Yi},
  booktitle={Proceedings of the 32nd ACM International Conference on Multimedia},
  pages={3877--3886},
  year={2024}
}

@article{xiao2024towards,
  title={Towards Visual Grounding: A Survey},
  author={Xiao, Linhui and Yang, Xiaoshan and Lan, Xiangyuan and Wang, Yaowei and Xu, Changsheng},
  journal={arXiv preprint arXiv:2412.20206},
  year={2024}
}

@article{chen2023shikra,
  title={Shikra: Unleashing multimodal llm's referential dialogue magic},
  author={Chen, Keqin and Zhang, Zhao and Zeng, Weili and Zhang, Richong and Zhu, Feng and Zhao, Rui},
  journal={arXiv preprint arXiv:2306.15195},
  year={2023}
}

@article{you2023ferret,
  title={Ferret: Refer and ground anything anywhere at any granularity},
  author={You, Haoxuan and Zhang, Haotian and Gan, Zhe and Du, Xianzhi and Zhang, Bowen and Wang, Zirui and Cao, Liangliang and Chang, Shih-Fu and Yang, Yinfei},
  journal={arXiv preprint arXiv:2310.07704},
  year={2023}
}

@inproceedings{wu2024v,
  title={V?: Guided visual search as a core mechanism in multimodal llms},
  author={Wu, Penghao and Xie, Saining},
  booktitle={Proceedings of the IEEE/CVF Conference on Computer Vision and Pattern Recognition},
  pages={13084--13094},
  year={2024}
}

@article{jiang2024marvel,
  title={Marvel: Multidimensional abstraction and reasoning through visual evaluation and learning},
  author={Jiang, Yifan and Sun, Kexuan and Sourati, Zhivar and Ahrabian, Kian and Ma, Kaixin and Ilievski, Filip and Pujara, Jay and others},
  journal={Advances in Neural Information Processing Systems},
  volume={37},
  pages={46567--46592},
  year={2024}
}

@article{sarch2024vlm,
  title={Vlm agents generate their own memories: Distilling experience into embodied programs of thought},
  author={Sarch, Gabriel and Jang, Lawrence and Tarr, Michael and Cohen, William W and Marino, Kenneth and Fragkiadaki, Katerina},
  journal={Advances in Neural Information Processing Systems},
  volume={37},
  pages={75942--75985},
  year={2024}
}

@article{guo2025deepseek,
  title={Deepseek-r1: Incentivizing reasoning capability in llms via reinforcement learning},
  author={Guo, Daya and Yang, Dejian and Zhang, Haowei and Song, Junxiao and Zhang, Ruoyu and Xu, Runxin and Zhu, Qihao and Ma, Shirong and Wang, Peiyi and Bi, Xiao and others},
  journal={arXiv preprint arXiv:2501.12948},
  year={2025}
}

@article{huang2025vision,
  title={Vision-r1: Incentivizing reasoning capability in multimodal large language models},
  author={Huang, Wenxuan and Jia, Bohan and Zhai, Zijie and Cao, Shaosheng and Ye, Zheyu and Zhao, Fei and Hu, Yao and Lin, Shaohui},
  journal={arXiv preprint arXiv:2503.06749},
  year={2025}
}

@article{shen2025vlm,
  title={Vlm-r1: A stable and generalizable r1-style large vision-language model},
  author={Shen, Haozhan and Liu, Peng and Li, Jingcheng and Fang, Chunxin and Ma, Yibo and Liao, Jiajia and Shen, Qiaoli and Zhang, Zilun and Zhao, Kangjia and Zhang, Qianqian and others},
  journal={arXiv preprint arXiv:2504.07615},
  year={2025}
}

@article{liu2025visual,
  title={Visual-rft: Visual reinforcement fine-tuning},
  author={Liu, Ziyu and Sun, Zeyi and Zang, Yuhang and Dong, Xiaoyi and Cao, Yuhang and Duan, Haodong and Lin, Dahua and Wang, Jiaqi},
  journal={arXiv preprint arXiv:2503.01785},
  year={2025}
}

@inproceedings{peng2024synthesize,
  title={Synthesize diagnose and optimize: Towards fine-grained vision-language understanding},
  author={Peng, Wujian and Xie, Sicheng and You, Zuyao and Lan, Shiyi and Wu, Zuxuan},
  booktitle={Proceedings of the IEEE/CVF Conference on Computer Vision and Pattern Recognition},
  pages={13279--13288},
  year={2024}
}

@article{zang2025contextual,
  title={Contextual object detection with multimodal large language models},
  author={Zang, Yuhang and Li, Wei and Han, Jun and Zhou, Kaiyang and Loy, Chen Change},
  journal={International Journal of Computer Vision},
  volume={133},
  number={2},
  pages={825--843},
  year={2025},
  publisher={Springer}
}

@inproceedings{yao2025map,
  title={MAP: Parameter-Efficient Tuning for Referring Expression Comprehension via Multi-Modal Adaptive Positional Encoding},
  author={Yao, Ruilin and Rong, Yi and Zou, Tianyu and Zhang, Bo and Li, Jian and Xiong, Shengwu and Xiong, Shili},
  booktitle={Proceedings of the 33rd ACM International Conference on Multimedia},
  pages={2264--2273},
  year={2025}
}

@inproceedings{wang2024haloquest,
  title={Haloquest: A visual hallucination dataset for advancing multimodal reasoning},
  author={Wang, Zhecan and Bingham, Garrett and Yu, Adams Wei and Le, Quoc V and Luong, Thang and Ghiasi, Golnaz},
  booktitle={European Conference on Computer Vision},
  pages={288--304},
  year={2024},
  organization={Springer}
}

@inproceedings{tong2024eyes,
  title={Eyes wide shut? exploring the visual shortcomings of multimodal llms},
  author={Tong, Shengbang and Liu, Zhuang and Zhai, Yuexiang and Ma, Yi and LeCun, Yann and Xie, Saining},
  booktitle={Proceedings of the IEEE/CVF Conference on Computer Vision and Pattern Recognition},
  pages={9568--9578},
  year={2024}
}

@inproceedings{wang2024all,
  title={The all-seeing project v2: Towards general relation comprehension of the open world},
  author={Wang, Weiyun and Ren, Yiming and Luo, Haowen and Li, Tiantong and Yan, Chenxiang and Chen, Zhe and Wang, Wenhai and Li, Qingyun and Lu, Lewei and Zhu, Xizhou and others},
  booktitle={European Conference on Computer Vision},
  pages={471--490},
  year={2024},
  organization={Springer}
}

@article{shao2024visual,
  title={Visual cot: Advancing multi-modal language models with a comprehensive dataset and benchmark for chain-of-thought reasoning},
  author={Shao, Hao and Qian, Shengju and Xiao, Han and Song, Guanglu and Zong, Zhuofan and Wang, Letian and Liu, Yu and Li, Hongsheng},
  journal={Advances in Neural Information Processing Systems},
  volume={37},
  pages={8612--8642},
  year={2024}
}

@article{Tian_Ma_Xie_Ye_2025, title={ChatterBox: Multimodal Referring and Grounding with Chain-of-Questions}, volume={39}, url={https://ojs.aaai.org/index.php/AAAI/article/view/32796}, DOI={10.1609/aaai.v39i7.32796}, abstractNote={In this study, we establish a benchmark and a baseline approach for Multimodal referring and grounding with Chain-of-Questions (MCQ), opening up a promising direction for ‘logical’ multimodal dialogues. The newly collected dataset, named CB-300K, spans challenges including probing dialogues with spatial relationship among multiple objects, consistent reasoning, and complex question chains. The baseline approach, termed ChatterBox, involves a modularized design and a referent feedback mechanism to ensure logical coherence in continuous referring and grounding tasks. This design reduces the risk of referential confusion, simplifies the training process, and presents validity in retaining the language model’s generation ability. Experiments show that ChatterBox demonstrates superiority in MCQ both quantitatively and qualitatively, paving a new path towards multimodal dialogue scenarios with logical interactions.}, number={7}, journal={Proceedings of the AAAI Conference on Artificial Intelligence}, author={Tian, Yunjie and Ma, Tianren and Xie, Lingxi and Ye, Qixiang}, year={2025}, month={Apr.}, pages={7401-7409} }

@article{xu2024mc,
  title={Mc-bench: A benchmark for multi-context visual grounding in the era of mllms},
  author={Xu, Yunqiu and Zhu, Linchao and Yang, Yi},
  journal={arXiv preprint arXiv:2410.12332},
  year={2024}
}

@article{fan2025scaling,
  title={Scaling Language-Free Visual Representation Learning},
  author={Fan, David and Tong, Shengbang and Zhu, Jiachen and Sinha, Koustuv and Liu, Zhuang and Chen, Xinlei and Rabbat, Michael and Ballas, Nicolas and LeCun, Yann and Bar, Amir and others},
  journal={arXiv preprint arXiv:2504.01017},
  year={2025}
}

@article{zhu2025internvl3,
  title={InternVL3: Exploring Advanced Training and Test-Time Recipes for Open-Source Multimodal Models},
  author={Zhu, Jinguo and Wang, Weiyun and Chen, Zhe and Liu, Zhaoyang and Ye, Shenglong and Gu, Lixin and Duan, Yuchen and Tian, Hao and Su, Weijie and Shao, Jie and others},
  journal={arXiv preprint arXiv:2504.10479},
  year={2025}
}

@article{bai2025qwen2,
  title={Qwen2. 5-vl technical report},
  author={Bai, Shuai and Chen, Keqin and Liu, Xuejing and Wang, Jialin and Ge, Wenbin and Song, Sibo and Dang, Kai and Wang, Peng and Wang, Shijie and Tang, Jun and others},
  journal={arXiv preprint arXiv:2502.13923},
  year={2025}
}

@article{chen2024more,
  title={Are more llm calls all you need? towards the scaling properties of compound ai systems},
  author={Chen, Lingjiao and Davis, Jared Quincy and Hanin, Boris and Bailis, Peter and Stoica, Ion and Zaharia, Matei A and Zou, James Y},
  journal={Advances in Neural Information Processing Systems},
  volume={37},
  pages={45767--45790},
  year={2024}
}

@article{wu2024deepseek,
  title={Deepseek-vl2: Mixture-of-experts vision-language models for advanced multimodal understanding},
  author={Wu, Zhiyu and Chen, Xiaokang and Pan, Zizheng and Liu, Xingchao and Liu, Wen and Dai, Damai and Gao, Huazuo and Ma, Yiyang and Wu, Chengyue and Wang, Bingxuan and others},
  journal={arXiv preprint arXiv:2412.10302},
  year={2024}
}

@article{team2024gemini,
  title={Gemini 1.5: Unlocking multimodal understanding across millions of tokens of context},
  author={Team, Gemini and Georgiev, Petko and Lei, Ving Ian and Burnell, Ryan and Bai, Libin and Gulati, Anmol and Tanzer, Garrett and Vincent, Damien and Pan, Zhufeng and Wang, Shibo and others},
  journal={arXiv preprint arXiv:2403.05530},
  year={2024}
}

@article{team2025gemma,
  title={Gemma 3 technical report},
  author={Team, Gemma and Kamath, Aishwarya and Ferret, Johan and Pathak, Shreya and Vieillard, Nino and Merhej, Ramona and Perrin, Sarah and Matejovicova, Tatiana and Ram{\'e}, Alexandre and Rivi{\`e}re, Morgane and others},
  journal={arXiv preprint arXiv:2503.19786},
  year={2025}
}

@inproceedings{lin2014microsoft,
  title={Microsoft coco: Common objects in context},
  author={Lin, Tsung-Yi and Maire, Michael and Belongie, Serge and Hays, James and Perona, Pietro and Ramanan, Deva and Doll{\'a}r, Piotr and Zitnick, C Lawrence},
  booktitle={Computer vision--ECCV 2014: 13th European conference, zurich, Switzerland, September 6-12, 2014, proceedings, part v 13},
  pages={740--755},
  year={2014},
  organization={Springer}
}

@article{everingham2010pascal,
  title={The pascal visual object classes (voc) challenge},
  author={Everingham, Mark and Van Gool, Luc and Williams, Christopher KI and Winn, John and Zisserman, Andrew},
  journal={International journal of computer vision},
  volume={88},
  pages={303--338},
  year={2010},
  publisher={Springer}
}

@article{fu2024mme,
  title={Mme-survey: A comprehensive survey on evaluation of multimodal llms},
  author={Fu, Chaoyou and Zhang, Yi-Fan and Yin, Shukang and Li, Bo and Fang, Xinyu and Zhao, Sirui and Duan, Haodong and Sun, Xing and Liu, Ziwei and Wang, Liang and others},
  journal={arXiv preprint arXiv:2411.15296},
  year={2024}
}

@article{li2024survey,
  title={A survey on multimodal benchmarks: In the era of large ai models},
  author={Li, Lin and Chen, Guikun and Shi, Hanrong and Xiao, Jun and Chen, Long},
  journal={arXiv preprint arXiv:2409.18142},
  year={2024}
}

@article{li2024embodied,
  title={Embodied agent interface: Benchmarking llms for embodied decision making},
  author={Li, Manling and Zhao, Shiyu and Wang, Qineng and Wang, Kangrui and Zhou, Yu and Srivastava, Sanjana and Gokmen, Cem and Lee, Tony and Li, Erran Li and Zhang, Ruohan and others},
  journal={Advances in Neural Information Processing Systems},
  volume={37},
  pages={100428--100534},
  year={2024}
}

@inproceedings{yue2024mmmu,
  title={Mmmu: A massive multi-discipline multimodal understanding and reasoning benchmark for expert agi},
  author={Yue, Xiang and Ni, Yuansheng and Zhang, Kai and Zheng, Tianyu and Liu, Ruoqi and Zhang, Ge and Stevens, Samuel and Jiang, Dongfu and Ren, Weiming and Sun, Yuxuan and others},
  booktitle={Proceedings of the IEEE/CVF Conference on Computer Vision and Pattern Recognition},
  pages={9556--9567},
  year={2024}
}

@inproceedings{liu2024mmbench,
  title={Mmbench: Is your multi-modal model an all-around player?},
  author={Liu, Yuan and Duan, Haodong and Zhang, Yuanhan and Li, Bo and Zhang, Songyang and Zhao, Wangbo and Yuan, Yike and Wang, Jiaqi and He, Conghui and Liu, Ziwei and others},
  booktitle={European conference on computer vision},
  pages={216--233},
  year={2024},
  organization={Springer}
}

@article{russakovsky2015imagenet,
  title={Imagenet large scale visual recognition challenge},
  author={Russakovsky, Olga and Deng, Jia and Su, Hao and Krause, Jonathan and Satheesh, Sanjeev and Ma, Sean and Huang, Zhiheng and Karpathy, Andrej and Khosla, Aditya and Bernstein, Michael and others},
  journal={International journal of computer vision},
  volume={115},
  pages={211--252},
  year={2015},
  publisher={Springer}
}

@article{achiam2023gpt,
  title={Gpt-4 technical report},
  author={Achiam, Josh and Adler, Steven and Agarwal, Sandhini and Ahmad, Lama and Akkaya, Ilge and Aleman, Florencia Leoni and Almeida, Diogo and Altenschmidt, Janko and Altman, Sam and Anadkat, Shyamal and others},
  journal={arXiv preprint arXiv:2303.08774},
  year={2023}
}

@article{zhang2024q,
  title={Q-bench: A benchmark for multi-modal foundation models on low-level vision from single images to pairs},
  author={Zhang, Zicheng and Wu, Haoning and Zhang, Erli and Zhai, Guangtao and Lin, Weisi},
  journal={IEEE Transactions on Pattern Analysis and Machine Intelligence},
  year={2024},
  publisher={IEEE}
}

@article{zhao2024octopus,
  title={Octopus: A multi-modal llm with parallel recognition and sequential understanding},
  author={Zhao, Chuyang and Song, YuXin and Chen, Junru and Rong, Kang and Feng, Haocheng and Zhang, Gang and Ji, Shufan and Wang, Jingdong and Ding, Errui and Sun, Yifan},
  journal={Advances in Neural Information Processing Systems},
  volume={37},
  pages={90009--90029},
  year={2024}
}

@article{wang2023large,
  title={Large language models are not fair evaluators},
  author={Wang, Peiyi and Li, Lei and Chen, Liang and Cai, Zefan and Zhu, Dawei and Lin, Binghuai and Cao, Yunbo and Liu, Qi and Liu, Tianyu and Sui, Zhifang},
  journal={arXiv preprint arXiv:2305.17926},
  year={2023}
}

@article{glm2024chatglm,
  title={Chatglm: A family of large language models from glm-130b to glm-4 all tools},
  author={GLM, Team and Zeng, Aohan and Xu, Bin and Wang, Bowen and Zhang, Chenhui and Yin, Da and Zhang, Dan and Rojas, Diego and Feng, Guanyu and Zhao, Hanlin and others},
  journal={arXiv preprint arXiv:2406.12793},
  year={2024}
}

@inproceedings{kirillov2023segment,
  title={Segment anything},
  author={Kirillov, Alexander and Mintun, Eric and Ravi, Nikhila and Mao, Hanzi and Rolland, Chloe and Gustafson, Laura and Xiao, Tete and Whitehead, Spencer and Berg, Alexander C and Lo, Wan-Yen and others},
  booktitle={Proceedings of the IEEE/CVF international conference on computer vision},
  pages={4015--4026},
  year={2023}
}

@article{liu2025bag,
  title={Bag of Tricks for Inference-time Computation of LLM Reasoning},
  author={Liu, Fan and Chao, Wenshuo and Tan, Naiqiang and Liu, Hao},
  journal={arXiv preprint arXiv:2502.07191},
  year={2025}
}

@article{kuznetsova2020open,
  title={The open images dataset v4: Unified image classification, object detection, and visual relationship detection at scale},
  author={Kuznetsova, Alina and Rom, Hassan and Alldrin, Neil and Uijlings, Jasper and Krasin, Ivan and Pont-Tuset, Jordi and Kamali, Shahab and Popov, Stefan and Malloci, Matteo and Kolesnikov, Alexander and others},
  journal={International journal of computer vision},
  volume={128},
  number={7},
  pages={1956--1981},
  year={2020},
  publisher={Springer}
}

@inproceedings{caesar2018coco,
  title={Coco-stuff: Thing and stuff classes in context},
  author={Caesar, Holger and Uijlings, Jasper and Ferrari, Vittorio},
  booktitle={Proceedings of the IEEE conference on computer vision and pattern recognition},
  pages={1209--1218},
  year={2018}
}

@article{young2014image,
  title={From image descriptions to visual denotations: New similarity metrics for semantic inference over event descriptions},
  author={Young, Peter and Lai, Alice and Hodosh, Micah and Hockenmaier, Julia},
  journal={Transactions of the association for computational linguistics},
  volume={2},
  pages={67--78},
  year={2014},
  publisher={MIT Press One Rogers Street, Cambridge, MA 02142-1209, USA journals-info~…}
}

@inproceedings{plummer2015flickr30k,
  title={Flickr30k entities: Collecting region-to-phrase correspondences for richer image-to-sentence models},
  author={Plummer, Bryan A and Wang, Liwei and Cervantes, Chris M and Caicedo, Juan C and Hockenmaier, Julia and Lazebnik, Svetlana},
  booktitle={Proceedings of the IEEE international conference on computer vision},
  pages={2641--2649},
  year={2015}
}

@article{schuhmann2022laion,
  title={Laion-5b: An open large-scale dataset for training next generation image-text models},
  author={Schuhmann, Christoph and Beaumont, Romain and Vencu, Richard and Gordon, Cade and Wightman, Ross and Cherti, Mehdi and Coombes, Theo and Katta, Aarush and Mullis, Clayton and Wortsman, Mitchell and others},
  journal={Advances in neural information processing systems},
  volume={35},
  pages={25278--25294},
  year={2022}
}

@article{madaan2023self,
  title={Self-refine: Iterative refinement with self-feedback},
  author={Madaan, Aman and Tandon, Niket and Gupta, Prakhar and Hallinan, Skyler and Gao, Luyu and Wiegreffe, Sarah and Alon, Uri and Dziri, Nouha and Prabhumoye, Shrimai and Yang, Yiming and others},
  journal={Advances in Neural Information Processing Systems},
  volume={36},
  pages={46534--46594},
  year={2023}
}

@inproceedings{
wang2025toolgen,
title={ToolGen: Unified Tool Retrieval and Calling via Generation},
author={Renxi Wang and Xudong Han and Lei Ji and Shu Wang and Timothy Baldwin and Haonan Li},
booktitle={The Thirteenth International Conference on Learning Representations},
year={2025},
url={https://openreview.net/forum?id=XLMAMmowdY}
}

@inproceedings{
gao2025multimodal,
title={Multi-modal Agent Tuning: Building a {VLM}-Driven Agent for Efficient Tool Usage},
author={Zhi Gao and Bofei Zhang and Pengxiang Li and Xiaojian Ma and Tao Yuan and Yue Fan and Yuwei Wu and Yunde Jia and Song-Chun Zhu and Qing Li},
booktitle={The Thirteenth International Conference on Learning Representations},
year={2025},
url={https://openreview.net/forum?id=0bmGL4q7vJ}
}

@inproceedings{
liu2025toolace,
title={Tool{ACE}: Winning the Points of {LLM} Function Calling},
author={Weiwen Liu and Xu Huang and Xingshan Zeng and xinlong hao and Shuai Yu and Dexun Li and Shuai Wang and Weinan Gan and Zhengying Liu and Yuanqing Yu and Zezhong WANG and Yuxian Wang and Wu Ning and Yutai Hou and Bin Wang and Chuhan Wu and Wang Xinzhi and Yong Liu and Yasheng Wang and Duyu Tang and Dandan Tu and Lifeng Shang and Xin Jiang and Ruiming Tang and Defu Lian and Qun Liu and Enhong Chen},
booktitle={The Thirteenth International Conference on Learning Representations},
year={2025},
url={https://openreview.net/forum?id=8EB8k6DdCU}
}

@article{zhang2024vipact,
  title={VipAct: Visual-perception enhancement via specialized vlm agent collaboration and tool-use},
  author={Zhang, Zhehao and Rossi, Ryan and Yu, Tong and Dernoncourt, Franck and Zhang, Ruiyi and Gu, Jiuxiang and Kim, Sungchul and Chen, Xiang and Wang, Zichao and Lipka, Nedim},
  journal={arXiv preprint arXiv:2410.16400},
  year={2024}
}

@inproceedings{li2023super,
  title={Super-clevr: A virtual benchmark to diagnose domain robustness in visual reasoning},
  author={Li, Zhuowan and Wang, Xingrui and Stengel-Eskin, Elias and Kortylewski, Adam and Ma, Wufei and Van Durme, Benjamin and Yuille, Alan L},
  booktitle={Proceedings of the IEEE/CVF conference on computer vision and pattern recognition},
  pages={14963--14973},
  year={2023}
}

@inproceedings{johnson2017clevr,
  title={Clevr: A diagnostic dataset for compositional language and elementary visual reasoning},
  author={Johnson, Justin and Hariharan, Bharath and Van Der Maaten, Laurens and Fei-Fei, Li and Lawrence Zitnick, C and Girshick, Ross},
  booktitle={Proceedings of the IEEE conference on computer vision and pattern recognition},
  pages={2901--2910},
  year={2017}
}

@article{liu2023visual,
  title={Visual spatial reasoning},
  author={Liu, Fangyu and Emerson, Guy and Collier, Nigel},
  journal={Transactions of the Association for Computational Linguistics},
  volume={11},
  pages={635--651},
  year={2023},
  publisher={MIT Press One Broadway, 12th Floor, Cambridge, Massachusetts 02142, USA~…}
}

@article{snell2024scaling,
  title={Scaling llm test-time compute optimally can be more effective than scaling model parameters},
  author={Snell, Charlie and Lee, Jaehoon and Xu, Kelvin and Kumar, Aviral},
  journal={arXiv preprint arXiv:2408.03314},
  year={2024}
}

@article{jaech2024openai,
  title={Openai o1 system card},
  author={Jaech, Aaron and Kalai, Adam and Lerer, Adam and Richardson, Adam and El-Kishky, Ahmed and Low, Aiden and Helyar, Alec and Madry, Aleksander and Beutel, Alex and Carney, Alex and others},
  journal={arXiv preprint arXiv:2412.16720},
  year={2024}
}

@article{team2025kimi,
  title={Kimi-vl technical report},
  author={Team, Kimi and Du, Angang and Yin, Bohong and Xing, Bowei and Qu, Bowen and Wang, Bowen and Chen, Cheng and Zhang, Chenlin and Du, Chenzhuang and Wei, Chu and others},
  journal={arXiv preprint arXiv:2504.07491},
  year={2025}
}

@inproceedings{deng2021transvg,
  title={Transvg: End-to-end visual grounding with transformers},
  author={Deng, Jiajun and Yang, Zhengyuan and Chen, Tianlang and Zhou, Wengang and Li, Houqiang},
  booktitle={Proceedings of the IEEE/CVF International Conference on Computer Vision},
  pages={1769--1779},
  year={2021}
}

@article{zong2024mova,
  title={Mova: Adapting mixture of vision experts to multimodal context},
  author={Zong, Zhuofan and Ma, Bingqi and Shen, Dazhong and Song, Guanglu and Shao, Hao and Jiang, Dongzhi and Li, Hongsheng and Liu, Yu},
  journal={Advances in Neural Information Processing Systems},
  volume={37},
  pages={103305--103333},
  year={2024}
}

@inproceedings{yang2022improving,
  title={Improving visual grounding with visual-linguistic verification and iterative reasoning},
  author={Yang, Li and Xu, Yan and Yuan, Chunfeng and Liu, Wei and Li, Bing and Hu, Weiming},
  booktitle={Proceedings of the IEEE/CVF Conference on Computer Vision and Pattern Recognition},
  pages={9499--9508},
  year={2022}
}

@article{xiao2023clip,
  title={Clip-vg: Self-paced curriculum adapting of clip for visual grounding},
  author={Xiao, Linhui and Yang, Xiaoshan and Peng, Fang and Yan, Ming and Wang, Yaowei and Xu, Changsheng},
  journal={IEEE Transactions on Multimedia},
  volume={26},
  pages={4334--4347},
  year={2023},
  publisher={IEEE}
}

@article{hong2025glm,
  title={GLM-4.1 V-Thinking: Towards Versatile Multimodal Reasoning with Scalable Reinforcement Learning},
  author={Hong, Wenyi and Yu, Wenmeng and Gu, Xiaotao and Wang, Guo and Gan, Guobing and Tang, Haomiao and Cheng, Jiale and Qi, Ji and Ji, Junhui and Pan, Lihang and others},
  journal={arXiv preprint arXiv:2507.01006},
  year={2025}
}

@inproceedings{chen2024efficient,
  title={An Efficient and Effective Transformer Decoder-Based Framework for Multi-task Visual Grounding},
  author={Chen, Wei and Chen, Long and Wu, Yu},
  booktitle={European Conference on Computer Vision},
  pages={125--141},
  year={2024}
}

@inproceedings{
you2024ferret,
title={Ferret: Refer and Ground Anything Anywhere at Any Granularity},
author={Haoxuan You and Haotian Zhang and Zhe Gan and Xianzhi Du and Bowen Zhang and Zirui Wang and Liangliang Cao and Shih-Fu Chang and Yinfei Yang},
booktitle={The Twelfth International Conference on Learning Representations},
year={2024},
url={https://openreview.net/forum?id=2msbbX3ydD}
}

@inproceedings{ma2024groma,
  title={Groma: Localized visual tokenization for grounding multimodal large language models},
  author={Ma, Chuofan and Jiang, Yi and Wu, Jiannan and Yuan, Zehuan and Qi, Xiaojuan},
  booktitle={European Conference on Computer Vision},
  pages={417--435},
  year={2024},
  organization={Springer}
}

@inproceedings{liu2025grounding,
  title={Grounding dino: Marrying dino with grounded pre-training for open-set object detection},
  author={Liu, Shilong and Zeng, Zhaoyang and Ren, Tianhe and Li, Feng and Zhang, Hao and others},
  booktitle={European Conference on Computer Vision},
  pages={38--55},
  year={2024},
  organization={Springer}
}

@article{dai2024simvg,
  title={Simvg: A simple framework for visual grounding with decoupled multi-modal fusion},
  author={Dai, Ming and Yang, Lingfeng and Xu, Yihao and Feng, Zhenhua and Yang, Wankou},
  journal={Advances in neural information processing systems},
  volume={37},
  pages={121670--121698},
  year={2024}
}

@article{zhang2025mllms,
  title={Mllms know where to look: Training-free perception of small visual details with multimodal llms},
  author={Zhang, Jiarui and Khayatkhoei, Mahyar and Chhikara, Prateek and Ilievski, Filip},
  journal={arXiv preprint arXiv:2502.17422},
  year={2025}
}

@inproceedings{shen2025zoomeye,
  title={Zoomeye: Enhancing multimodal llms with human-like zooming capabilities through tree-based image exploration},
  author={Shen, Haozhan and Zhao, Kangjia and Zhao, Tiancheng and Xu, Ruochen and Zhang, Zilun and Zhu, Mingwei and Yin, Jianwei},
  booktitle={Proceedings of the 2025 Conference on Empirical Methods in Natural Language Processing},
  pages={6613--6629},
  year={2025}
}

@article{zhang2025mllms_s,
  title={Why do mllms struggle with spatial understanding? a systematic analysis from data to architecture},
  author={Zhang, Wanyue and Huang, Yibin and Xu, Yangbin and Huang, JingJing and Zhi, Helu and Ren, Shuo and Xu, Wang and Zhang, Jiajun},
  journal={arXiv preprint arXiv:2509.02359},
  year={2025}
}

@article{azzolini2025cosmos,
  title={Cosmos-reason1: From physical common sense to embodied reasoning},
  author={Azzolini, Alisson and Bai, Junjie and Brandon, Hannah and Cao, Jiaxin and Chattopadhyay, Prithvijit and Chen, Huayu and Chu, Jinju and Cui, Yin and Diamond, Jenna and Ding, Yifan and others},
  journal={arXiv preprint arXiv:2503.15558},
  year={2025}
}

@inproceedings{gupta2023visual,
  title={Visual programming: Compositional visual reasoning without training},
  author={Gupta, Tanmay and Kembhavi, Aniruddha},
  booktitle={Proceedings of the IEEE/CVF conference on computer vision and pattern recognition},
  pages={14953--14962},
  year={2023}
}

@article{chen2023autoagents,
  title={Autoagents: A framework for automatic agent generation},
  author={Chen, Guangyao and Dong, Siwei and Shu, Yu and Zhang, Ge and Sesay, Jaward and Karlsson, B{\"o}rje F and Fu, Jie and Shi, Yemin},
  journal={arXiv preprint arXiv:2309.17288},
  year={2023}
}

@article{huang2024understanding,
  title={Understanding the planning of LLM agents: A survey},
  author={Huang, Xu and Liu, Weiwen and Chen, Xiaolong and Wang, Xingmei and Wang, Hao and Lian, Defu and Wang, Yasheng and Tang, Ruiming and Chen, Enhong},
  journal={arXiv preprint arXiv:2402.02716},
  year={2024}
}

@inproceedings{zhang2024timearena,
  title={TimeArena: Shaping Efficient Multitasking Language Agents in a Time-Aware Simulation},
  author={Zhang, Yikai and Yuan, Siyu and Hu, Caiyu and Richardson, Kyle and Xiao, Yanghua and Chen, Jiangjie},
  booktitle={Proceedings of the 62nd Annual Meeting of the Association for Computational Linguistics (Volume 1: Long Papers)},
  pages={3894--3916},
  year={2024}
}

@inproceedings{yuan2025easytool,
  title={Easytool: Enhancing llm-based agents with concise tool instruction},
  author={Yuan, Siyu and Song, Kaitao and Chen, Jiangjie and Tan, Xu and Shen, Yongliang and Ren, Kan and Li, Dongsheng and Yang, Deqing},
  booktitle={Proceedings of the 2025 Conference of the Nations of the Americas Chapter of the Association for Computational Linguistics: Human Language Technologies (Volume 1: Long Papers)},
  pages={951--972},
  year={2025}
}

@article{zhang2025survey,
  title={A survey on the memory mechanism of large language model-based agents},
  author={Zhang, Zeyu and Dai, Quanyu and Bo, Xiaohe and Ma, Chen and Li, Rui and Chen, Xu and Zhu, Jieming and Dong, Zhenhua and Wen, Ji-Rong},
  journal={ACM Transactions on Information Systems},
  volume={43},
  number={6},
  pages={1--47},
  year={2025},
  publisher={ACM New York, NY}
}

@article{chen2024persona,
  title={From persona to personalization: A survey on role-playing language agents},
  author={Chen, Jiangjie and Wang, Xintao and Xu, Rui and Yuan, Siyu and Zhang, Yikai and Shi, Wei and Xie, Jian and Li, Shuang and Yang, Ruihan and Zhu, Tinghui and others},
  journal={arXiv preprint arXiv:2404.18231},
  year={2024}
}

@inproceedings{hong2023metagpt,
  title={MetaGPT: Meta programming for a multi-agent collaborative framework},
  author={Hong, Sirui and Zhuge, Mingchen and Chen, Jonathan and Zheng, Xiawu and Cheng, Yuheng and Wang, Jinlin and Zhang, Ceyao and Wang, Zili and Yau, Steven Ka Shing and Lin, Zijuan and others},
  booktitle={The Twelfth International Conference on Learning Representations},
  year={2023}
}

@inproceedings{chen2023agentverse,
  title={Agentverse: Facilitating multi-agent collaboration and exploring emergent behaviors},
  author={Chen, Weize and Su, Yusheng and Zuo, Jingwei and Yang, Cheng and Yuan, Chenfei and Chan, Chi-Min and Yu, Heyang and Lu, Yaxi and Hung, Yi-Hsin and Qian, Chen and others},
  booktitle={The Twelfth International Conference on Learning Representations},
  year={2023}
}
	\bibliographystyle{iclr2026_conference}
	\newpage
	\appendix
	\section{Appendix}
	\subsection{Related work} \label{related_work}
	\subsubsection{Benchmarks for Visual Capability of MLLMs}
	The capability of Visual Perception, Understanding and Reasoning is a foundational aspect of
	understanding benchmarks, which involves the ability to recognize and localize multiple objects, interpret various visual elements with complex emotional or implicit cues and summarize visual information for feedback and decision making \citep{li2024embodied}.
	Specifically, Perception in MLLMs involves the classification, detection of basic visual objects (\textit{e.g.}, dog, cat) and attributes (\textit{e.g.}, color, lighting). These low-level
	perceptual capabilities are crucial for various applications,
	including recognition systems \citep{zhao2024octopus} and visual quality enhancement \citep{zhang2024q}. Understanding represents
	a sophisticated level of image understanding
	that focuses on the detailed and nuanced aspects of
	visual content. It includes recognizing and interpreting
	the visual-linguistic concepts, such as text recognition (OCRBench \citep{liu2024ocrbench}), Visual Grounding (RefCOCO \citep{yu2016modeling}, FineCops-Ref \citep{liu2024finecops}, HC-RefLoCo \citep{wei2024large}) and Referring Expression Generation (Visual Genome) \citep{krishna2017visual}, which refers to the model’s ability to accurately link visual
	elements with corresponding textual descriptions. Although tasks at this level begin to involve visual and textual alignment, they still do not require reasoning or external knowledge. For higher-order capability, reasoning in MLLMs involves advanced event understanding and deep meaning extraction from multimodal data. These capabilities include interpreting and responding to complex emotional cues across multiple modalities \citep{cheng2024emotion}, deriving subtle implicit meanings from visual and contextual information \citep{liu2023multilingual}, and a range of other competencies, including knowledge acquisition, language generation, spatial awareness, and cultural context integration \citep{rachabatuni2024context}.
	\subsubsection{Reasoning Capability of MLLMs} MLLMs have demonstrated remarkable reasoning capabilities, largely facilitated by test-time scaling \citep{dong2022survey, wei2022chain}, which allows feeding prompted samples and
	context. This capability has been further enhanced by chain-of-thought (CoT) prompting \citep{wei2022chain}, which
	enables LLMs to generate coherent intermediate reasoning steps toward the final answer. Previous
	studies have shown that LLMs benefit from manually written demonstrations as well as zero-shot
	prompting outputs. However, due to the domain gap between various modalities, the current reasoning capability of MLLMs in the complex real-world environment is still limited. 
	To address this limitation, researchers have focused
	on enhancing the reasoning capability of MLLMs in both the training and prompting paradigms. Flamingo \citep{alayrac2022flamingo} bridges the gap between these two modalities by pre-training on interleaved
	visual and textual data. Some other works, such as Shikra \citep{chen2023shikra} and Ferret \citep{you2023ferret}, leverage visual grounding data \citep{xiao2024towards, yao2024visual} to achieve fine-grained vision-language alignment. 
	Furthermore, recent studies have also demonstrated that augmenting computing resources during the testing phase (test-time scaling) can enhance the reasoning capabilities of LLMs \citep{jaech2024openai}. More specifically, Prompt-based Reasoning Meta-Systems (PRMS) can be employed to guide LLMs in evaluating and filtering intermediate ``thinking'' processes \citep{snell2024scaling}. This encourages the generation of more sophisticated reasoning steps during testing, ultimately leading to improved reasoning accuracy.
	Beyond that, some methods employ the external knowledge to focus on important visual details, like V*~\citep{wu2024v}, Marvel \citep{jiang2024marvel}, and ICAL 
	\citep{sarch2024vlm}, collecting a series of visual reasoning steps as training data.
	More recently, with the emergence of DeepSeek-R1 \citep{guo2025deepseek} demonstrating strong potential in LLM reasoning, research efforts have begun to explore reasoning-centric models and R1-style reinforcement learning strategies for understanding complex visual scenes and tasks. These studies \citep{huang2025vision,shen2025vlm,liu2025visual} particularly emphasize the long-chain reasoning capabilities within MLLMs, aiming to enhance their performance in handling intricate visual-linguistic reasoning challenges.

    \subsubsection{LLM-Based Agentic Reasoning} 
    The rapid progress of large language models \citep{achiam2023gpt, bai2025qwen2} has sparked significant interest in building autonomous agents capable of solving complex, multi-step reasoning tasks. Leveraging the strong chain-of-thought (CoT) abilities of modern LLMs \citep{wei2022chain}, these systems typically decompose a complex problem into a sequence of structured subtasks, invoke intermediate deliberation, and integrate the resulting insights to produce a final answer \citep{gupta2023visual, chen2023autoagents}. Recent developments in LLM-based autonomous agents highlight the importance of planning \citep{huang2024understanding, zhang2024timearena}, tool usage \citep{yuan2025easytool}, memory \citep{zhang2025survey}, and persona \citep{chen2024persona}. In parallel, multi-agent frameworks such as MetaGPT \citep{hong2023metagpt}, AgentVerse \citep{chen2023agentverse}, etc., demonstrate strong performance by orchestrating multiple interacting agents, often instantiated as distinct roles with specialized responsibilities. Despite their success, these systems rely heavily on manually designed personas, fixed role hierarchies, or hand-crafted coordination rules, which limits their flexibility and generalization across tasks and domains. Moreover, the dependence on external scaffolding or pre-specified agent behaviors often restricts the model’s ability to adaptively adjust its internal reasoning pathway. To address these limitations, SMEC introduces a self-driven agentic reasoning mechanism that automatically generates diverse experts, elicits their reasoning, filters redundant experts, and synthesizes their perspectives into a final consensus, all within the model’s own language-native inference loop.
	
	\subsection{Specific defination of different tasks} \label{fig:specific_defination}
	\textbf{Object Counting \textbf(OC):} Estimating the number of object instances described by a free-form expression, often under complex conditions like occlusion, scale variation, or clutter.  
	
	\textbf{Object Detection (OD):} Localizing objects within an image by generating bounding boxes paired with corresponding class labels. In order to better match the real-life scenarios and practical applications, we construct more than 500 fine-grained object categories based on natural language.
	
	\textbf{Object Existence Determination (OE):} Determining whether a particular object, which described by a detailed expression, exists in the image without requiring spatial localization. 
	
	\textbf{Relation Extraction (RE):} Identifying semantic relationships (\textit{e.g.}, ``holding'', ``next to'', ``wearing'') between pairs of objects to facilitate structured scene understanding. And we added questions about the objects that do not exist in the images to evaluate model's ability to suppress hallucinations.
	
	\textbf{Visual Grounding (VG):} Localizing an image region that corresponds to a natural language expression, linking linguistic references to fine-grained visual content. 
	
	\textbf{Region-wise OCR (OCR): }Recognizing and transcribing text within a region, which specified by coordinates or description, facilitating fine-grained interleaved image-text understanding.
	
	\textbf{Spatial Relationship Comprehension (SRC):} Understanding geometric relationships (\textit{e.g.}, ``above'' and ``to the left front to'') between objects within diverse 3D views, supporting visual-spatial reasoning. Compared to some rudimentary or synthetic spatial understanding datasets \citep{johnson2017clevr, li2023super, liu2023visual}, our data is more realistic in emphasizing spatial location understanding under real-world scenarios as well as 2D images acquired by cameras or cell phones.
	
	\textbf{Scene Knowledge Inference (SKI):} Inferring high-level semantic and functional information about the scene or making decision based on the visual contents, incorporating context, commonsense knowledge, and visual cues beyond explicit visual entities. Compared to the regular visual reasoning dataset, \data{} additionally distinguish between ``thought paths'' and ``final answers'', differentiated by the $<think>$ token, aiming to provide finer-grained information for potential test-time scaling tests and R1-style reinforcement learning.
    	\begin{figure}[h]
		\centering
		\includegraphics[width=0.89\linewidth]{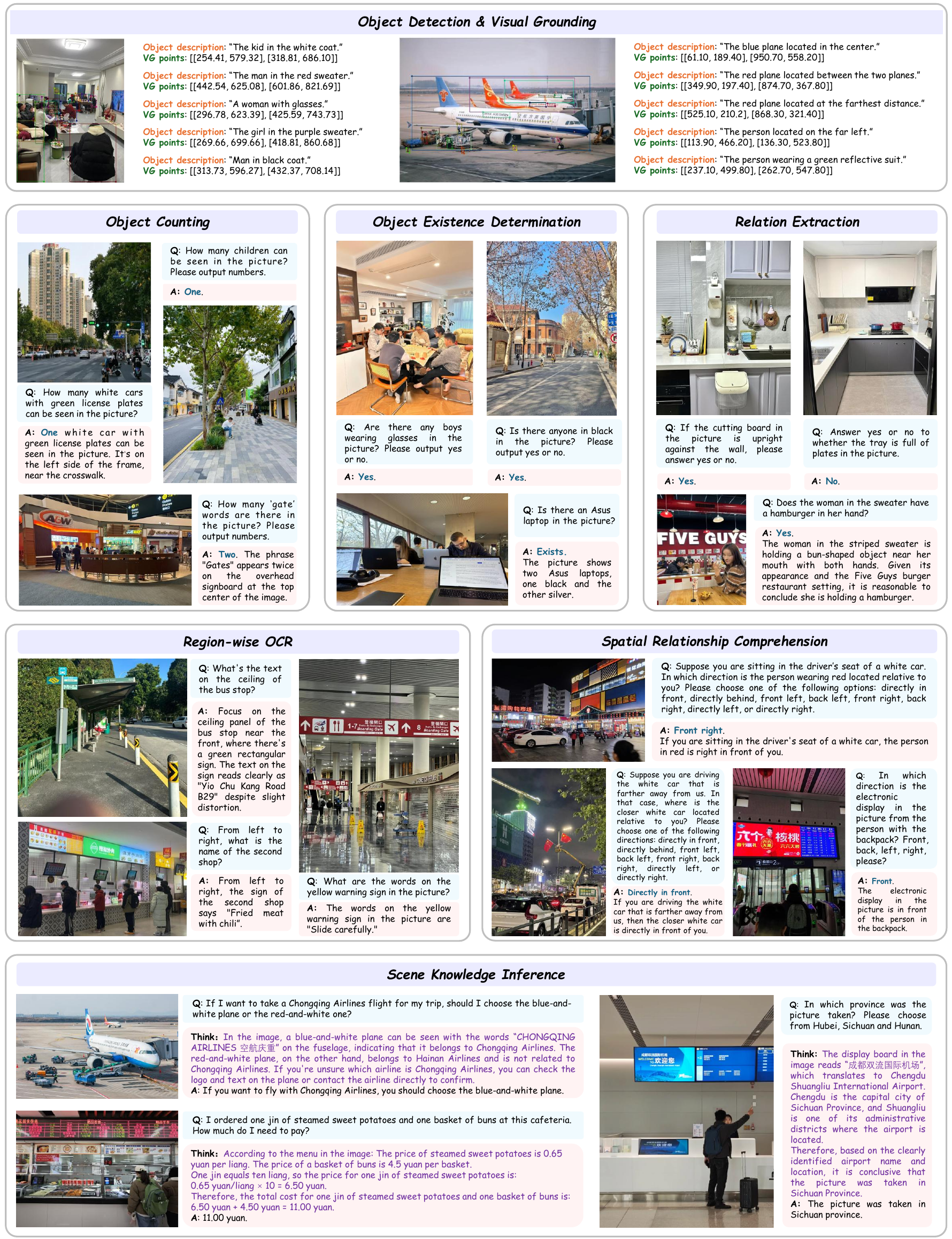}\vspace{-7pt}
		\caption{
			\data{} covers a wide range of images and annotations, from fine-grained recognition and spatial localization to complex reasoning over extended thought processes. Notably, each image is annotated with labels corresponding to all subtasks concurrently, enabling comprehensive evaluation.
		}\vspace{-27pt}
		\label{fig:samples}
	\end{figure}
	\subsection{Quality control process}\label{fig:Quality}
	In addition to the annotations, diversified measures were taken to enrich the content of data samples and ensure their quality. Specifically, we implemented a multi-faceted quality control process. Beyond a two-step data cleaning protocol, we also enriched each image with supplementary metadata to facilitate traceability and contextual analysis.

First, we manually performed a two-stage data cleaning process. In the initial stage, we reviewed and eliminated suspected duplicate images. The second stage involved distributing the problems among co-authors for meticulous format and typo checking, ensuring all annotations adhered to a standardized format.

To further validate the quality and consistency of our annotations, we performed an additional two-step verification process. This included both manual and machine-assisted checks. The entire dataset was cross-verified by both an independent team of annotators and the Qwen2.5-VL 72B open-source model. For machine validation, we input the original image, question, and answer into the MLLM. Cases flagged as invalid by the model were isolated for manual re-evaluation by a separate team of annotators. For object detection and visual grounding tasks, we directly visualized the annotations on the images, enabling human evaluators to assess the validity of the bounding boxes.

Additionally, we enriched each image with supplementary metadata. We included a pseudonymized Annotator ID to allow for annotator-specific quality tracking while preserving privacy. The Time of Online Publication and Scene Category were also labeled to facilitate temporal studies, filter outdated content, and organize the dataset by scene. Finally, ambiguous images that consistently resulted in low inter-annotator agreement were manually filtered out to ensure a high-quality final dataset. Some cleaned representative examples are visualized in Figure \ref{fig:samples}.
\subsection{Data Privacy Protection and Copyright Statement}\label{sec:Privacy}
Our protocol for handling potentially sensitive information was a multi-stage process designed to be as thorough as possible:

\textbf{Automated Pre-screening:} As an initial step, we used automated tools (\textit{e.g.}, face detection models \footnote{https://github.com/timesler/facenet-pytorch}
, docTR \footnote{https://github.com/mindee/doctr}) to perform a preliminary scan of the collected images. This scan was configured to flag images with a high probability of containing human faces or dense blocks of text that might constitute personally identifiable information.

\textbf{Comprehensive Manual Review:} Every image, including those not flagged by the automated scan, was then subjected to a thorough manual review by our team of over 20 trained human annotators. Annotators received specific training and a detailed guide on identifying a wide range of sensitive data, including but not limited to:
Visible and recognizable faces; Full names, usernames, or contact information; License plates, street addresses, or other specific location markers; Private documents or screens displaying personal data.

\textbf{Sensitive Information Exclusion:} Images containing sensitive personal information were either excluded or processed to blur or mask sensitive regions, to mitigate privacy risks. Based on the manual review, if an image contained sensitive information, one of two actions was taken as mentioned in the paper:

\begin{itemize}
\item Processing: If the sensitive information was incidental to the image's main content, we applied irreversible blurring or masking to the specific region.
\item Exclusion: If the sensitive information was central to the image and could not be adequately anonymized without destroying the scene's context, the image was entirely excluded from the final dataset.
\end{itemize}

\textbf{Final Verification:} To ensure consistency and quality, a final audit was conducted by a subset of the paper's authors. This team reviewed a random sample of the approved images and 100\% of the processed (blurred/masked) images to verify that our privacy protocol was correctly and consistently applied.

This multi-stage, human-centric approach ensures that the images in the \data{} dataset comply with platform policies  and respect individual privacy. The \data{} dataset is constructed solely for non-commercial academic research purposes and does not transfer copyright ownership of the original images. By accessing or using the \data{} dataset, users agree to comply with all applicable copyright laws, platform policies, and research-use restrictions.

	\begin{figure*}[t]
		\centering
		\includegraphics[width=\linewidth]{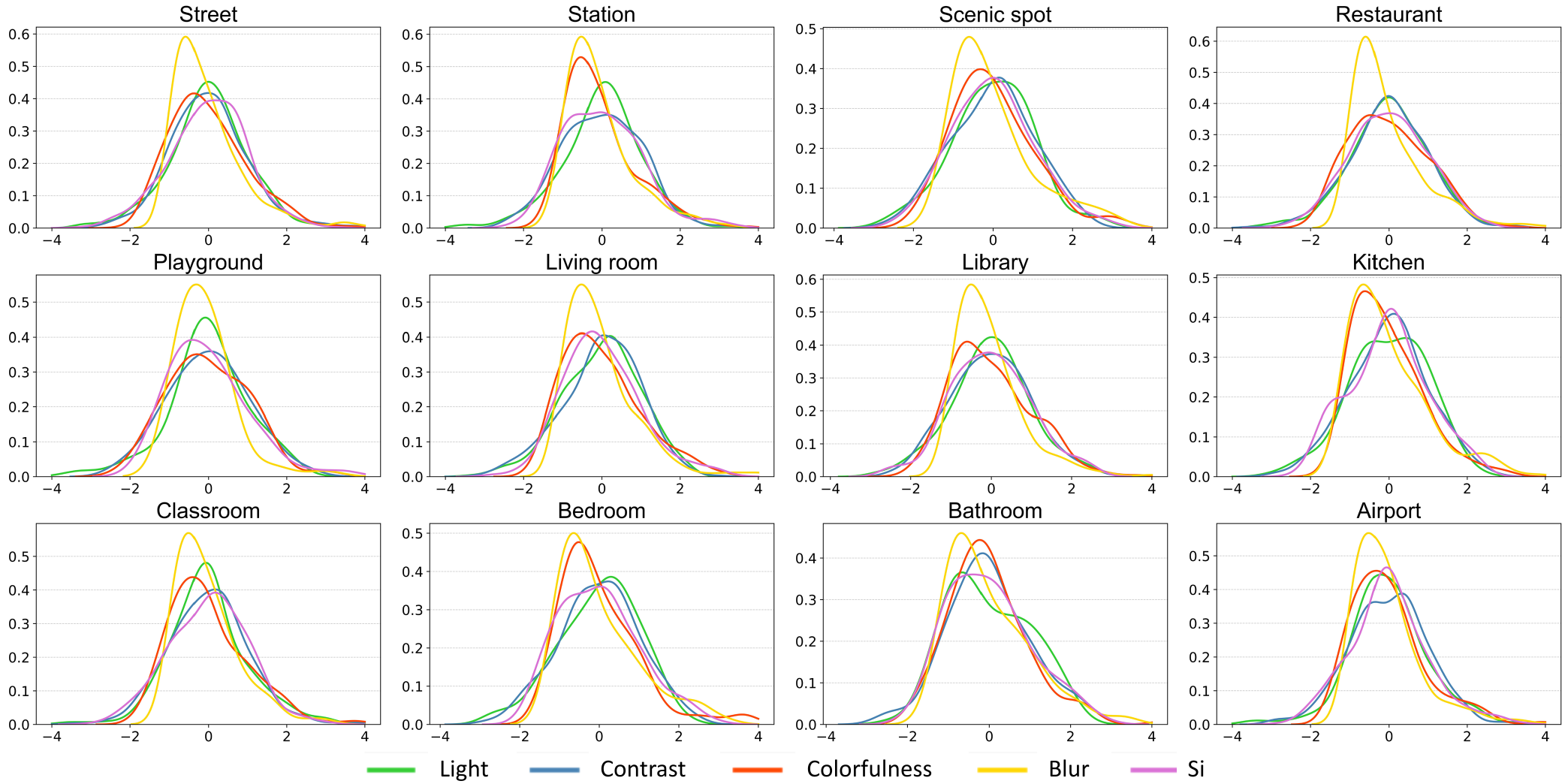}
		\caption{The normalized probability distributions of low-level attributes from different scenes. Scenes with flat peaks show more diversity, while those with sharp peaks have similar features.}
		\label{fig:lowlevelanalysis}
	\end{figure*}
    
	\subsection{Low-level feature analysis of images from different scenes}\label{low_level}
	We counted five low-level visual attributes, including lighting, contrast, color, blur, and spatial information (SI), to assess the statistical difference between different scenes.  As shown in Figure  \ref{fig:lowlevelanalysis}, the normalized probability density curves of low-level visual attributes across different scenes are consistent with human perceptual preferences. Scenes with regulated lighting conditions (\textit{e.g.}, classrooms, airports, and stations) demonstrate sharp peaks near x $\approx$ 0 in the illumination curves (density $>$ 0.5), indicating constrained variations in brightness. In contrast, domestic environments (\textit{e.g.}, living rooms, bedrooms, and kitchens) display broader illumination distributions, suggesting more diverse and adaptive light sources. Furthermore, functional scenes such as bedrooms, bathrooms, and kitchens exhibit sharp, concentrated peaks in color distributions (peak density $\approx$ 0.5), implying greater structural regularity in specific visual attributes.

\subsection{Fine-grained evaluation of VG} 
    We conducted a fine-grained evaluation of a diverse set of models on the Visual Grounding (VG) task \cite{you2024ferret, bai2025qwen2, yao2025map, zhu2025internvl3, xiao2023clip}, categorizing them into two groups: traditional predictive multimodal models and generative multimodal large language models (MLLMs). The results, summarized in Table \ref{tab:rec_sota}, reveal several key insights into the current state of visual grounding capabilities.

\subsubsection{Comparison of Model Categories}

The results clearly indicate a significant performance gap between the two model categories. The top-performing MLLMs, specifically Qwen2.5-VL-7B and Qwen2.5-VL-32B, demonstrate superior performance across all metrics, with an accuracy of 46.94\% and 48.47\% at $IoU@0.5$, respectively. This performance is substantially higher than the top predictive model, G-DINO, which achieves 37.05\% at the same metric. This finding suggests that the generative and in-context learning capabilities of modern MLLMs provide a substantial advantage in the complex VG task, enabling them to better understand nuanced linguistic instructions and ground them accurately in the visual space.

\subsubsection{Performance on Different Scales}

A multi-scale analysis, measured by accuracy on small ($ACC_{s}$), medium ($ACC_{m}$), and large ($ACC_{l}$) objects, provides a more granular view of each model's strengths and weaknesses. Both traditional and generative models exhibit a similar trend: performance consistently improves with the size of the target object. This is a common challenge in visual grounding and object detection, as localizing and grounding small objects remains difficult.

 \textbf{Traditional Models:} Among the traditional models, G-DINO demonstrates a more balanced performance across scales, achieving 24.87\% on small objects and 52.32\% on large objects. In contrast, models like VLTVG and SimVG struggle significantly with small objects, with accuracies of 0.00\% and 0.01\% respectively, but show strong performance on large objects (29.70\% and 45.20\%).

 \textbf{Generative MLLMs:} While MLLMs also struggle with small objects, their performance is notably better than most traditional models. Qwen2.5-VL-32B and Qwen2.5-VL-7B achieve high accuracy on medium and large objects, with their $ACC_{m}$
  and $ACC_{l}$
  scores reaching 55.04\% and 61.48\% (for Qwen2.5-VL-32B), and 54.12\% and 60.24\% (for Qwen2.5-VL-7B) respectively. The strong performance on larger objects may be attributed to their powerful visual backbones and advanced language understanding capabilities, which help them better contextualize the target within the scene.

\subsubsection{Impact of Model Architecture and Size}

Our results also highlight the importance of model architecture and size. The Qwen2.5-VL family of models, with its impressive performance, benefits from a powerful visual encoder (FE-ViT) and a sophisticated Qwen2.5 language backbone. Similarly, the InternVL3 series shows a clear scaling effect, where performance on most metrics improves as the model size increases from 2B to 14B. The performance of the 38B variant is slightly lower than the 14B variant due to its different visual backbone. This trend, consistent with findings in large language models, suggests that scaling up both visual and linguistic components is a promising direction for future research in visual grounding.
    \begin{table*}[!ht]
	\footnotesize
	\centering
	\resizebox{1.0\textwidth}{!}{%
		\begin{tabular}{l|c|c|ccccc|ccc}
			\toprule
			\multirow{2}{*}{Method}  & Visual   & Linguistic  & \multicolumn{5}{c|}{Accuracy @ IoU} & \multicolumn{3}{c}{Scale-wise Accuracy}  \\
			 & Backbone & Backbone  &   $@0.5$  & $@0.6$ & $@0.7$  &   $@0.8$  & $@0.9$ & $ACC_{s}$  & $ACC_{m}$& $ACC_{l}$ \\
			\midrule    
			\multicolumn{11}{l}{\textbf{Methods based on predictive multimodal models: }}   \\ 
            \midrule    
			TransVG \citep{deng2021transvg}       & RN101 & BERT-B  & 8.73&7.57&6.29&4.40&1.69&0.01&2.01&23.64\\
VLTVG~\citep{yang2022improving}    & RN101 & BERT-B   & 11.04&9.60&7.75&5.33&1.99&0.00&2.80&29.70\\
MMCA \citep{yao2024visual}& RN101& BERT-B & 10.92&9.45&7.90&5.64&2.22&0.03 &2.79&29.31\\
CLIP-VG \citep{xiao2023clip}     & CLIP-B & CLIP-B    & 8.73 & 7.57 & 6.29& 4.40&1.69& 0.01 & 2.01 & 23.64\\ 
    EEVG~\citep{chen2024efficient}    & ViT-B/16 & BERT-B  & 9.27 & 5.78 & 2.51 & 0.48 & 0.05&0.01 & 0.98 & 26.12 \\ 
            SimVG \citep{dai2024simvg}  & BEIT-3 & BEIT-3&16.46 &13.90 &11.12 &7.44 &2.70 &0.01 &3.10 &45.20\\
            G-DINO \citep{liu2025grounding}  & Swin-L & BERT-B   & 37.05 &33.92 &29.20 &22.57 &11.36 &24.87 &37.54 &52.32\\
			\midrule  
			\multicolumn{11}{l}{\textbf{Methods based on generative multimodal large language models: }}      \\ 
            \midrule    
            Groma-7B \citep{ma2024groma}  & DINOv2-L &  Vicuna  & 33.59& 29.95 &25.47 &18.73 &8.52 &11.58 &33.91 &58.59 \\
            Mova-7B \citep{zong2024mova}& Multi-expert& Vicuna &20.44 &13.10 &5.98 &1.09 &0.13 &5.06 &15.97 &40.36\\
            Ferret-7B \citep{you2024ferret} & CLIP-L & Vicuna & 23.26& 18.97 &13.95 &7.61 &1.84 &1.85 &19.49 &54.64\\
            Ferret-13B \citep{you2024ferret} & CLIP-L & Vicuna & 24.20 &19.81 &14.41 &8.12 &2.05 &2.26 &20.42 &56.31\\
            InternVL3-2B \citep{zhu2025internvl3}& InternViT-0.3B&Qwen2.5&7.89 &5.10 &2.85 &1.36 &0.33 &0.61 &3.46 &19.34\\
            InternVL3-8B \citep{zhu2025internvl3}& InternViT-0.3B&Qwen2.5&17.54 &13.23 &8.89 &4.94 &1.60 &3.23 &15.36 &35.21\\
            InternVL3-14B \citep{zhu2025internvl3} & InternViT-0.3B&Qwen2.5&29.53 &23.98 &17.25 &10.05 &3.00 &4.58 &27.80 &57.07\\
            InternVL3-38B \citep{zhu2025internvl3}&InternViT-6B&Qwen2.5&27.85&21.42 &15.00 &8.23 &2.37 &4.91 &25.81 &53.56\\
            VLM-R1-3B \citep{shen2025vlm} & FE-ViT&Qwen2.5 &23.79 &19.91 &15.65 &10.53 &4.31 &8.15 &22.84 &40.94\\
		Qwen2.5-VL-3B \citep{bai2025qwen2} & FE-ViT&Qwen2.5&45.03&37.92&29.33&18.48&6.51 &29.14&50.54&57.15\\
            Qwen2.5-VL-7B \citep{bai2025qwen2} & FE-ViT&Qwen2.5& 46.94 & 39.39&29.94&18.38&6.26&28.87&54.12&60.24\\
            Qwen2.5-VL-32B \citep{bai2025qwen2} & FE-ViT&Qwen2.5& 48.47 & 40.66&30.78&19.15&6.63&30.93&55.04&61.48\\	
			\bottomrule
		\end{tabular}%
	}
    \caption{Multi-scale evaluation results.}
	\label{tab:rec_sota}%
\end{table*}%
\subsection{More Synergistic Effects Analysis} \label{more_synergistic}
We further notice that Figure \ref{fig:fine_grained} (a) shows OC-OCR correlation (0.77) $\gg$ OC-OE (0.46). This contradicts intuition—object counting (OC) should align more naturally with existence checks (OE) than OCR. We attribute this to two factors. First, the Object Existence (OE) task is a simple binary classification: either an object is present or it is not. It requires a model to make a broad, scene-level assessment. In contrast, Object Counting (OC) is a more demanding task that requires fine-grained localization of individual objects, followed by an enumeration step. A model can be highly proficient at a binary existence check without possessing the precise localization and counting skills needed for the OC task. This fundamental difference in cognitive demand limits the correlation between the two. Second, the high correlation between Object Counting (OC) and OCR is not coincidental. Both tasks rely on a critical shared capability: fine-grained localization. To perform well on the OC task, a model must accurately identify and localize each instance of an object to count it. Similarly, to perform region-wise OCR, the model must first precisely locate the bounding box of the text before reading it. The strong correlation suggests that the ability to perform precise object localization is a dominant factor in a model's success on both tasks, thus strengthening their relationship despite their different end goals.

To further test the synergy between different tasks, we conducted a experiment with Qwen2.5-VL-7B on a sampled subset of \data{}. Specifically, when testing the SKI task, we fed the VQA question-answer pairs of other tasks into the model as context along with the question, and asked it to return the answer (refer to Appendix \ref{sec:p_synergy} for the prompt template $p_s$, where we provide an example based on the OCR task). The test results are shown in Table \ref{tab:synergistic_effect} and reveal several noteworthy patterns. First, incorporating OCR information yields a substantial performance gain (from 39.80\% to 41.36\%), indicating that understanding scene text helps the model solve some reasoning tasks. Second, although some tasks—such as OE and OC—exhibit limited or even negative effects when introduced individually, their combination with OCR consistently boosts performance. This may indicate that auxiliary perceptual signals, while insufficient on their own, can enhance reasoning when mediated through textual understanding. Furthermore, the observed synergistic effects resonate with the design philosophy of our proposed Self-Driven Multi-Expert Collaborative (SMEC) framework. We argue that complex multimodal reasoning cannot be achieved by isolated competencies alone, but requires a structured mechanism to coordinate heterogeneous sources of evidence. 

\begin{table*}[!ht]
	\footnotesize
	\centering
\begin{tabular}{c|c|c|c|c|c|c|c|c|c|c|c}
\toprule
  OCR &-& \ding{52} &- & -& - &\ding{52}&\ding{52}&\ding{52}&\ding{52}&\ding{52}&\ding{52}\\
  RE &- & - & \ding{52} & - &- & \ding{52}&-&-&\ding{52}&\ding{52}&\ding{52}\\
   OE &- & - & -& \ding{52} &-& -&\ding{52}& -&\ding{52}&-&\ding{52}\\
   OC &- & -&-&- & \ding{52} & -&-&\ding{52}&-&\ding{52}&\ding{52}\\
   \midrule
   Performance&39.80& 41.36&38.03&39.23 & 39.52& 40.72& 41.45& 40.73&41.93&40.79&41.90\\
\bottomrule
\end{tabular}
\caption{Testing the synergistic effects of different tasks on Scene Knowledge Inference (SKI).}
	\label{tab:synergistic_effect}%
\end{table*}%

    \subsection{Analysis of Input Resolution}
\begin{table*}[!ht]
	\footnotesize
	\centering
	\resizebox{1.0\textwidth}{!}{%
		\begin{tabular}{l|c|c|cccc}
			\toprule
			\multirow{2}{*}{Method}  & Visual   & Linguistic  & \multicolumn{4}{c}{Accuracy @ Input Resolution} \\
			 & Backbone & Backbone  & 640 $\times$ 640 & 960  $\times$ 960& 1280  $\times$ 1280 &   1600  $\times$ 1600 \\
            \midrule    
            InternVL3-2B \citep{zhu2025internvl3}& InternViT-0.3B&Qwen2.5& 40.97&41.53&41.90&40.60\\
            InternVL3-9B \citep{zhu2025internvl3}& InternViT-0.3B&InternLM3-8B& 46.95& 46.54& 46.65&46.63\\
            InternVL3-14B \citep{zhu2025internvl3} & InternViT-0.3B&Qwen2.5& 50.28 & 51.15& 51.58&51.17\\
            InternVL3-38B \citep{zhu2025internvl3}&InternViT-6B&Qwen2.5& 50.88&50.97&49.98&51.08\\
		Qwen2.5-VL-3B \citep{bai2025qwen2} & FE-ViT&Qwen2.5&40.08&40.44&40.48&40.58\\
            Qwen2.5-VL-7B \citep{bai2025qwen2} & FE-ViT&Qwen2.5& 46.52 & 47.41&48.13&48.11\\
            Qwen2.5-VL-32B \citep{bai2025qwen2} & FE-ViT&Qwen2.5& 53.72 & 54.37&54.10 &54.09\\
            GLM-4.1V-Base-9B \citep{hong2025glm}& AlMv2-Huge&GLM-4-0414& 42.35 & 42.86&43.28 &43.73\\
            GLM-4.1V-Thinking-9B \citep{hong2025glm} & AlMv2-Huge&GLM-4-0414& 48.77 & 50.78&51.32 &51.13\\
			\bottomrule
		\end{tabular}%
	}
    \caption{Benchmark results  across varying input resolutions.}\vspace{-10pt}
	\label{tab:vqa_sota}%
\end{table*}%
    We conducted a detailed analysis to understand the impact of varying input resolutions on model performance. The results, summarized in Table \ref{tab:vqa_sota}, reveal several key insights.

\textbf{General Trend (Performance Improves with Resolution):}
For most models, performance generally improves as the input resolution increases. This trend is evident in models such as Qwen2.5-VL-7B, which shows a steady increase in accuracy from 46.52\% at 640$\times$640 to 48.13\% at 1280x1280. Similarly, GLM-4.1V-Thinking-9B improves from 48.77\% to 51.32\% over the same range. This is expected, as higher resolutions provide more visual detail, which is particularly beneficial for complex visual grounding and reasoning tasks that require fine-grained perception.

\textbf{The Point of Diminishing Returns:}
However, the results also suggest a point of diminishing returns. For many models, the performance gain from increasing the resolution beyond 1280x1280 is minimal, and in some cases, performance slightly decreases. For example, InternVL3-14B peaks at 51.58\% at 1280x1280 and then slightly drops to 51.17\% at 1600x1600. Similarly, Qwen2.5-VL-7B's performance plateaus at 1280$\times$1280. This phenomenon could be attributed to several factors, including the model's architecture, which may not be fully optimized to handle the extra high-resolution information, or the fact that the added detail does not contribute meaningfully to solving the task.

\textbf{Model-Specific Variations:}
Interestingly, some models, like InternVL3-2B, show less sensitivity to resolution changes, with its performance remaining relatively stable across all resolutions. In contrast, models such as GLM-4.1V-Thinking-9B and Qwen2.5-VL-32B demonstrate a more pronounced performance improvement with higher resolutions, indicating that their architectures are more capable of leveraging the extra visual information. This suggests that the optimal input resolution is not a one-size-fits-all solution and depends heavily on the model's architecture and design.
\begin{figure}[t]\vspace{-2pt}
		\centering
		\includegraphics[width=\linewidth]{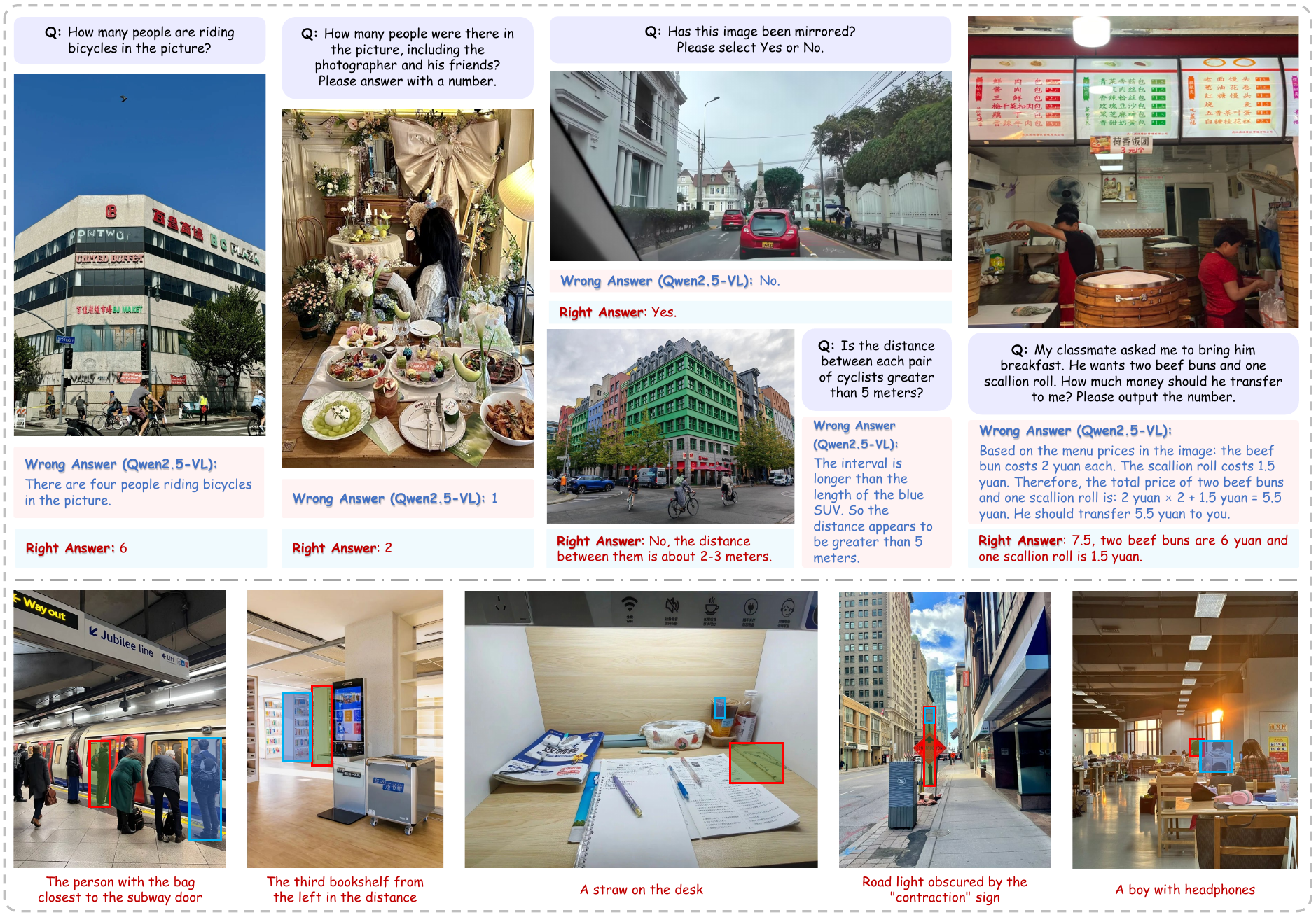}
		\caption{Failure cases for various tasks. Incorrect predictions and labels are indicated by blue and red, respectively}
		\label{fig:qualiative_erroe_results}\vspace{-10pt}
	\end{figure}
        \subsection{Qualitative error analysis}\vspace{-3pt}
     To better illustrate common failure patterns and the underlying limitations of current Multimodal Large Language Models, we conduct a qualitative analysis of representative error cases across different task levels. Following the structure of our benchmark, we group the visualizations into two categories: (1) VQA-style tasks and (2) Localization tasks, including detection and visual grounding, as shown in Figure \ref{fig:qualiative_erroe_results}. 

     For Perception \& Understanding Tasks, they primarily require directly aligning visual content with textual queries. While modern MLLMs achieve reasonably high accuracy, their errors frequently stem from low-level perceptual limitations such as small objects, occlusion, distant subjects, and sensitivity to resolution. Several examples clearly illustrate this issue—for instance, miscounting the number of cyclists in a street scene, or failing to detect objects like a straw on a desk or a road light partially obscured by signage. For more complex reasoning tasks, we observe two major classes of systematic errors.

\textbf{Correct reasoning, incorrect perception:} The model often demonstrates sound logical reasoning but bases its inference on incorrect or incomplete visual extraction. For example, in a price-computation task, although the model performs the arithmetic correctly, it misreads the menu price of the beef bun, leading to a wrong total (5.5 instead of 7.5 yuan). This reveals a persistent bottleneck where high-level reasoning is constrained by low-level perception, especially OCR and fine-grained attribute recognition. Such cases motivate the need for agent-based or expert-collaborative pipelines, which can iteratively refine visual cues or invoke specialized perception experts. They also highlight the importance of dynamic zoom-in strategies for capturing critical but small textual or visual details. \citep{wu2024v, zhang2025mllms, shen2025zoomeye, shao2024visual}

\textbf{Spatial and physical reasoning deficits:} A second recurring failure mode involves questions requiring geometric or physical commonsense. Models frequently struggle with tasks that implicitly require depth understanding, object-scale priors, or spatial metric reasoning. For instance, the model incorrectly concludes that two cyclists are more than 5 meters apart by comparing them to the length of a blue SUV, even though the correct distance is only about 2–3 meters. Likewise, it fails to judge whether an image is mirrored due to misunderstanding spatial layout cues. These issues echo recent findings showing that current MLLMs still lack robust spatial grounding and physical commonsense. \citep{zhang2025mllms_s, azzolini2025cosmos}

	\subsection{A Formal Description of SMEC} \label{algorithm1}
	As shown in Algorithm \ref{alg:algorithm}. A key advantage of SMEC is that it does not depend on fixed, hand-crafted prompts. Instead, the prompts and expert descriptions are self-generated by the model based on the given visual input and question. During each iteration, the model adaptively refines its expert descriptions and updates the generation process when redundancy or low-quality information is detected. This adaptive design means that SMEC is not tied to a specific phrasing or a predefined set of experts, but can flexibly adjust to different problems and question types. As a result, our method is more robust than approaches that rely heavily on manually designed prompts, since the “experts” in SMEC emerge dynamically from the model itself rather than being externally imposed.
	\begin{figure}[htb]
		\begin{center}\vspace{-10pt}
			\begin{minipage}{12cm}
				\begin{algorithm}[H]
					\caption{Self-driven Multi-Expert Generation \& Collaboration}
					\label{alg:algorithm}
					\textbf{Initialization}: Based Instruction-tuned MLLM $\theta$, Question $q$,  Meta Generation Prompt $p_g$, Inspection prompt $p_i$, Collaboration Prompt $p_c$, Description Set $D= \emptyset$, Maximum Answer Set $A= \emptyset$, Iterations $N_t$. 
					\begin{algorithmic}[1] 
						\STATE $a_0 = \theta(q), A = A \cup a_0$ \hfill $\#$ Initial answer for question.
						\FOR {$t=1,2,\ldots, N_t$}
						\IF {$t=1$} 
						\STATE $d_q^1 = \theta(p_g, q, A_0)$ \hfill $\#$ Initial expert description.
						\ELSE
						\STATE $d_q^t = \theta(p_g, D, q, A_t)$ \hfill $\#$ New description based on existing information.
						\ENDIF
						\IF {$\theta(p_i, D, d_q^t) = Retain $ } 
						\STATE $D = D \cup d_q^t $ \hfill $\#$ Checking process.
						\STATE  $a_t = \theta(q, d_q^t), A = A \cup a_t$ \hfill $\#$ New answer from the expert perspective.
						\ELSE 
						\STATE $p_g = \theta(q, d_q^t, p_g, D)$ \hfill $\#$ Update generation prompt while repeat descriptions.
						\ENDIF
						\ENDFOR 
						\STATE $a_{final} = \theta(q, A, p_c, D) $\hfill $\#$ Summarize the final answer.
					\end{algorithmic}
				\end{algorithm}
			\end{minipage}
		\end{center}\vspace{-10pt}
	\end{figure}\vspace{-10pt}
    \subsection{Human Preference}\label{sec:human} 
    To ensure the verifiability of our evaluation, particularly for open-ended reasoning tasks, we employed a large language model (LLM) as an automatic grader. To mitigate the concern regarding potential LLM hallucinations or failure to detect nuanced mistakes, it is noted that the LLM grader (\textit{e.g.}, GLM4-flash) is used to compare the model-generated responses against our pre-existing, human-annotated answers, ensuring that the ground truth remains anchored in high-quality human data. 
We also conducted a human preference analysis on a representative and complex task SKI with a subset of the dataset, aiming to provide a gold standard against which to measure the reliability of judgement model. We recruited a separate team of ten expert annotators who were not involved in the original data collection to manually evaluate the accuracy of the model's answers compared to the labeled answers, as shown in Table \ref{tab:preference}. In this setup, Actual Positive (AP) and Actual Negative (AN) represent human judgments of correctness and incorrectness, respectively, while Test Positive and Test Negative indicate the LLM grader’s corresponding evaluations.

The results, shown in Table \ref{tab:preference}, demonstrate a strong alignment between human preference and LLM-based judgments across all evaluated models. For instance, in the case of Qwen2.5-VL's responses, when humans labeled an answer as correct (AP), the LLM grader also marked it as correct 97.13\% of the time. Similarly, when humans judged an answer as incorrect (AN), the LLM grader agreed 96.14\% of the time. Comparable trends are observed for InternVL3 and Gemini2.5-pro, though with slightly larger gaps in negative cases. These findings suggest that the LLM grader provides a highly reliable approximation of human judgment, especially for positive cases. Incorporating human validation thus not only confirms the feasibility of using LLMs as evaluators but also highlights their potential to scale evaluation consistently across large datasets while retaining strong alignment with expert human preference.
\begin{table*}[!ht]
	\footnotesize
	\centering
\begin{tabular}{c|cc|cc|cc}
\toprule
\multirow{2}{*}{\textbf{Method}} &\multicolumn{2}{c|}{\textbf{Qwen2.5-VL}}&\multicolumn{2}{c|}{\textbf{InternVL3}}&\multicolumn{2}{c}{\textbf{Gemini2.5-pro}}\\
& \textbf{AP} & \textbf{AN} & \textbf{AP} & \textbf{AN} & \textbf{AP} & \textbf{AN}\\
\midrule
\textbf{Test Positive} & 97.13\% & 2.87\% &94.70\%&5.30\% & 93.65\% &6.35\%\\
\midrule
\textbf{Test Negative} & 3.86\% & 96.14\% & 7.77\%&92.23\% &11.41\% &88.59\%\\
\bottomrule
\end{tabular}
\caption{Human evaluation for the models' responses.}
	\label{tab:preference}\vspace{-10pt}
\end{table*}%
\begin{figure}[t]
		\centering
    \includegraphics[width=\linewidth]{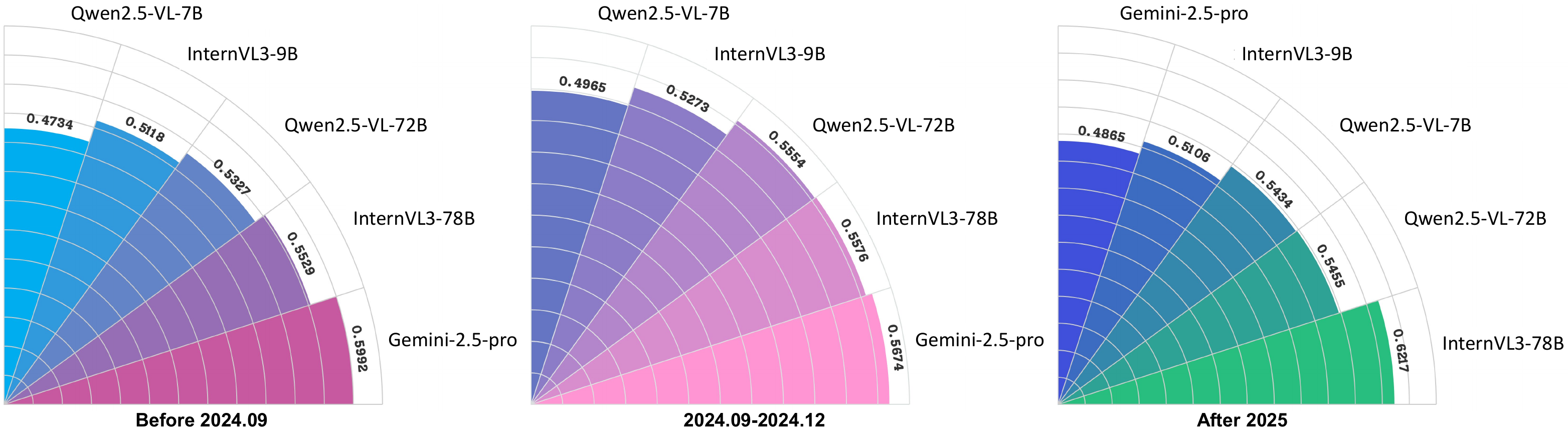}\vspace{-5pt}
		\caption{Model accuracy on SKI task across temporal split.}
		\label{fig:time_acc_ski}\vspace{-10pt}
	\end{figure}
\subsection{Temporal Generalization Performance}\vspace{-3pt}
We visualize the accuracy of the best few models on data from different time periods in Figure \ref{fig:time_acc_ski}, including Qwen2.5-VL (7B/72B), InternVL3 (9B/78B), and Gemini-2.5-Pro. Notably, models released after late 2024, such as InternVL3-78B and Qwen2.5-VL-72B, consistently outperform Gemini-2.5-pro on contemporary data, although their overall performance lower than Gemini-2.5-pro. The experimental results further supporting the observation that newer models tend to generalize better to new data distributions. This trend underscores the impact of scaling, instruction tuning, and exposure to temporally aligned data in enhancing multimodal reasoning performance.
    \begin{figure}[t]
		\centering
    \includegraphics[width=\linewidth]{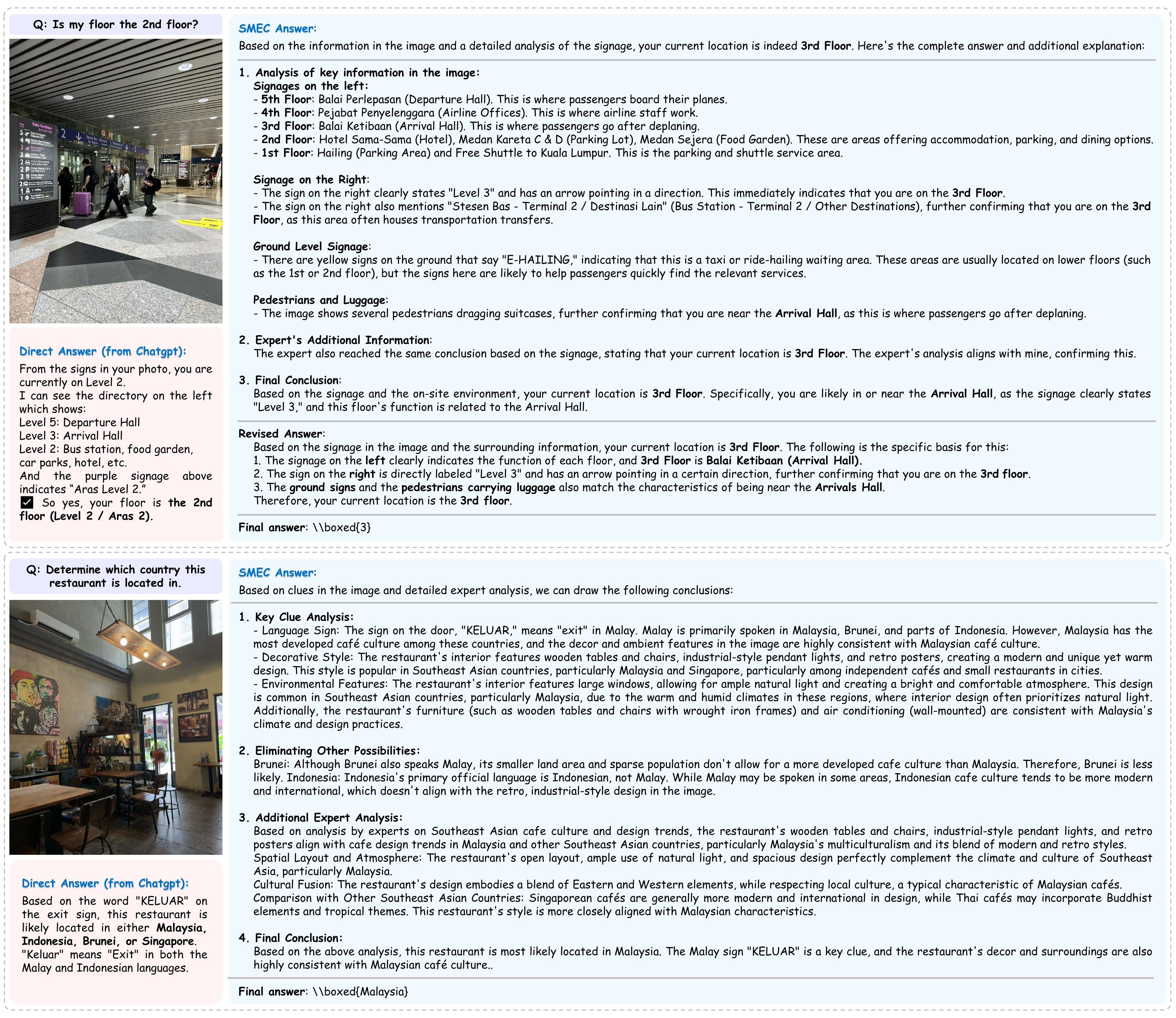}\vspace{-5pt}
		\caption{Comparison of our method and ChatGPT on some difficult examples.}
		\label{fig:failure_cases}\vspace{-15pt}
	\end{figure}
\subsection{More on SMEC vs. Baseline methods}
To further highlight the advantage of our proposed SMEC framework, we provide additional qualitative comparisons against baseline methods, including direct inference and ChatGPT-style single-pass reasoning. Figure \ref{fig:failure_cases} showcases representative challenging samples drawn from \data{}, where baseline models tend to produce either incomplete or overconfident predictions.

In these cases, ChatGPT and other baselines often failed for two recurring reasons:

\textbf{Over-Reliance on Surface Cues.} Baselines typically produced answers anchored on the most salient visual elements, neglecting contextual or relational signals. For instance, when asked to infer spatial constraints or traffic rules in a airport photo, ChatGPT tended to extrapolate directly from textual OCR cues, leading to plausible but incorrect answers.

\textbf{Lack of Internal Deliberation.} Without multi-perspective reasoning, baselines converged prematurely on a single hypothesis. This often caused brittle errors in scenarios requiring integration of textual, spatial, and commonsense evidence.

By contrast, SMEC decomposed the problem into multiple role-specific perspectives, such as a scene analyst, a spatial reasoner, and a cultural or commonsense expert. These experts generated partially overlapping but complementary hypotheses, which were then screened for redundancy and synthesized into a consensus. In the traffic-sign example shown in Figure \ref{fig:failure_cases}, SMEC correctly filtered out spurious cues and converged on the right driving instruction, whereas ChatGPT remained uncertain or hallucinated unsupported details. Across difficult samples, two consistent patterns emerged:

\textbf{Error Correction through Redundancy Filtering.} Even when some experts produced misleading interpretations, SMEC’s screening stage effectively down-weighted unreliable reasoning chains, preventing error propagation.

\textbf{Multi-Step Enrichment.} Iterative expert collaboration yielded richer reasoning trajectories, allowing the framework to exploit synergies between visual grounding, OCR, and commonsense inference. This process systematically improved robustness to ambiguous or noisy inputs.

Taken together, these results suggest that SMEC is not merely a test-time ensemble but a principled framework that encourages internal debate within a single model. Unlike single-pass inference, SMEC operationalizes a language-native form of deliberation, enabling models to approximate the dynamics of human expert panels. As Figure \ref{fig:failure_cases} illustrates, this mechanism directly translates into more accurate and interpretable reasoning on complex multimodal questions.
    \subsection{Prompts for Synergy Testing}\label{sec:p_synergy}
    \begin{tcolorbox}[title = {Prompt $p_s$ with OCR question-answer pairs as context}] 
		$\{question~from ~SKI\}$
		
		You can use the following facts to help you answer this question. Please note that they may not be relevant to the question. Here are some factual questions and answers about this picture:
		
		$\{question~from~OCR\}$: $\{answer~from~OCR\}$
	\end{tcolorbox}
	\subsection{Prompts of Self-driven Multi-Expert Collaborative Framework}
	\begin{tcolorbox}[title = {Meta Generation Prompt $p_g$}] 
		$\{question\}$
		
		Here are your answers and those of some experts:
		
		$\{answer\}$
		
		Now you can create and work with multiple experts to improve your answer. So, please describe in as much detail as possible the different skills and focus you need from each expert.
		
		We will provide each expert with the same information and queries. Each expert should have his or her own specialization covering perception, understanding and reasoning, etc., so you can assign only one subtask to each expert to ensure a more refined answer.
		We will relay their responses to you in turn so that you can reorganize them into better answers. Please note that descriptions should be in the second person, e.g. You are XXX.
		
		These are the descriptions of the experts you have previously created for this task:
		
		$\{description\}$
		
		Therefore, do not create the same experts as above over and over again.
		
		Now you can create a description for the new expert (please note that you can only describe one, not more than one at the same time):
	\end{tcolorbox}
	
	\begin{tcolorbox}[title = {Inspection Prompt $p_c$}] 
		$\{question\}$
		
		We hired multiple experts to answer this question. Below is a second person description of the experts we hired:
		$\{existing~description\}$
		
		We are now hiring a new expert to help better provide the information needed for the question as well as respond to user queries. Here is a second person description of the new expert:
		$\{description\}$
		
		Since there is an additional cost to hiring a new Expert, please evaluate the new Expert based on the following two criteria to decide whether or not to retain them. 
		
		1. based on the new Expert's description, determine if they can effectively assist in answering the user's question or provide the information needed for the question.
		
		2. the new expert is not a duplicate of any existing expert.
		
		The new expert must meet both of these criteria. If either criterion is not met, they should be discarded.
		If retaining, please reply `Retain'. If discarded, please reply: `Discard'.
	\end{tcolorbox}
	
	\begin{tcolorbox}[title = {Collaboration Prompt $p_c$}] 
		$\{question\}$
		These are you and some experts' answer:
		$\{answer\}$
		
		The description of the experts you invited are: $\{description\}$
		
		Now, you can refine your answer based on the answer and additional information they provided to better answer the question.
		Keep in mind that the experts' answer and additional information may not be correct, so decide carefully whether to accept his answer or stick to your original one.
        
		Revised answer:
	\end{tcolorbox}
	
	\subsection{Limitations} While \data{} offers broad task coverage and a unified evaluation setting, it currently focuses on static images and short-form reasoning. Real-world applications may require multimodal reasoning over temporal sequences or long-form narratives, which are beyond the scope of this version. Additionally, SMEC relies on prompt-based expert simulation, which, though flexible, may introduce redundancy or sensitivity to prompt phrasing and inference. 
\subsection{The Use of Large Language Models (LLMs)}
We have not used Large Language Models (LLMs) for our paper writing.	
	
\end{document}